\pdfoutput=1

\documentclass[11pt]{article}

\usepackage[final]{acl}

\usepackage{times}
\usepackage{latexsym}

\usepackage[T1]{fontenc}
\usepackage[utf8]{inputenc}

\usepackage{microtype}

\usepackage{inconsolata}

\usepackage{graphicx}

\title{

Evaluating Large Language Models for Detecting Antisemitism

}

\author{
 \textbf{Jay Patel\textsuperscript{1}},  \hspace{0.5em} 
 \textbf{Hrudayangam Mehta\textsuperscript{1}}, \hspace{0.5em} 
 \textbf{Jeremy Blackburn\textsuperscript{1}} \hspace{0.5em} 
\\
\texttt{\{jpatel67, hmehta, jblackbu\}@binghamton.edu}
\\
 \textsuperscript{1}Binghamton University, NY, USA
\\
}

\makeatletter
\newcommand{\appendixtableofcontents}{%
  \begingroup
    \setcounter{tocdepth}{-10}
    \tableofcontents
  \endgroup
}
\makeatother

\usepackage{enumitem}

\usepackage{tikz}
\usepackage{amsmath}
\usepackage{booktabs}

\usepackage{subcaption}

\usepackage[most]{tcolorbox}
\usepackage{xcolor}
\usepackage{tabularx}

\usepackage{pifont}

\usepackage[table]{xcolor}

\usepackage{changepage} %

\definecolor{limegreen}{RGB}{50,205,50}
\definecolor{darkgreen}{RGB}{0,100,0}

\newif\ifcomment
\commenttrue
\ifcomment
\newcommand{\redcolor}[1]{{\bf \textcolor{red}{#1}}}
\newcommand{\jbnote}[1]{{\bf \textcolor{magenta}{JB: #1}}}
\else
\newcommand{\redcolor}[1]{}
\newcommand{\jbnote}[1]{}
\fi

\begin{document}
\maketitle

\begin{abstract}

Detecting hateful content is a challenging and important problem.
Automated tools, like machine-learning models, can help, but they require continuous training to adapt to the ever-changing landscape of social media.
In this work, we evaluate eight open-source LLMs' capability to detect antisemitic content, specifically leveraging in-context definition.
We also study how LLMs understand and explain their decisions given a moderation policy as a guideline.
First, we explore various prompting techniques and design a new CoT-like prompt, Guided-CoT, and find that injecting domain-specific thoughts increases performance and utility.
Guided-CoT handles the in-context policy well, improving performance and utility by reducing refusals across all evaluated models, regardless of decoding configuration, model size, or reasoning capability.
Notably, Llama 3.1 70B outperforms fine-tuned GPT-3.5.
Additionally, we examine LLM errors and introduce metrics to quantify semantic divergence in model-generated rationales, revealing notable differences and paradoxical behaviors among LLMs.
Our experiments highlight the differences observed across LLMs' utility, explainability, and reliability.~\footnote{Code and resources available at: \url{https://github.com/idramalab/quantify-llm-explanations}}~\footnote{\textcolor{blue}{Accepted to EMNLP 2025 Main Conference}}

\end{abstract}

\maketitle

\section{Introduction}

\begin{figure}[t]
\centering
\includegraphics[width=0.70\linewidth]{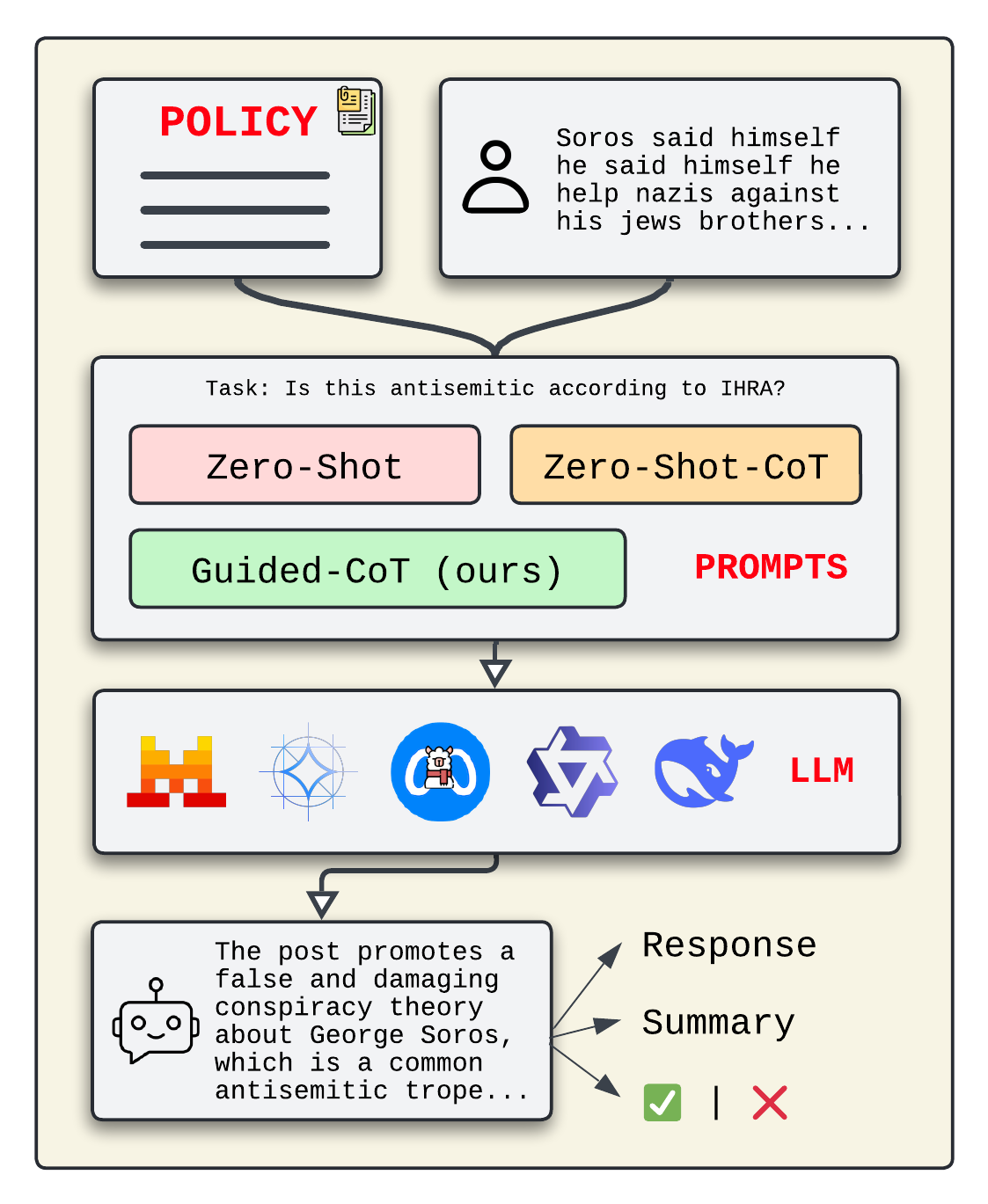}
\caption{Evaluation of open-source LLMs using an in-context policy-oriented guideline for the classification task of detecting antisemitism.
}
\label{fig:paper-teaser}
\end{figure}

\redcolor{Warning: The content in paper may be distressing or offensive for some readers.}

To combat hate speech (e.g., antisemitism), social media platforms moderate content according to a set of policies, but these moderation policies are complex, nuanced, and dependent on local laws and societal norms~\citep{common2020fear}.
This complexity makes developing and deploying automated systems, like various machine-learning models, challenging beyond technical efforts, and practical implementation involves interdisciplinary stakeholders.
For example, the International Holocaust Remembrance Alliance (IHRA) provides a starting point for identifying antisemitism and that has been adopted by many governments and institutions, including the U.S. Department of State~\citep{ihra-def, ihra-usa, ihra-members}.
Ultimately, moderation policies are a socio-technical specification.

Recent advances in large language models (LLMs) offer promising directions for moderating (e.g., detecting) harmful content online~\citep{roy2023probing, piot2024decoding, alkhamissi2022token, thapa2025large, yin2025bingoguard}.
~\citet{becker2024antisemitism} fine-tuned a proprietary GPT-3.5 model for antisemitism detection using a non-publicly available annotated dataset of about 22K samples.
However, fine-tuning has its challenges: it requires an annotated dataset, which is time-consuming, lacks a universal harm taxonomy, and lacks the scalability to keep up with the changing social media landscape~\citep{chen2024class, palla2025policy}.

Researchers have investigated alternatives to fine-tuning, like in-context prompting with reasoning approaches, e.g., Chain-of-Thought (CoT) and its adaptive variants, which aim to enhance reasoning by including multiple perspectives or abstract reasoning guides~\citep{just-etal-2025-dipt, liao-etal-2025-forest}.
However, LLMs' utility in sensitive tasks remains limited due to its safety-tuned release~\citep{vijjini-etal-2025-exploring, zhang2025falsereject}, raising concerns about fully utilizing LLMs for nuanced tasks like detecting antisemitism.
To explore this avenue further, we evaluate eight open-source LLMs to detect antisemitism via in-context learning without fine-tuning, e.g., explicit policy-oriented guideline (the IHRA definition with contemporary examples) in the prompt.

We explore multiple prompting techniques, including Zero-Shot, CoT, and a newly designed Guided-CoT, followed by an ablation study to identify more or less important in-context thoughts for our task.
Additionally, we examine how the models perform when policy is provided in the prompt, across different decoding strategies and prompting techniques.
Furthermore, we investigate the quantitative differences in LLM-generated explanations between content classified as antisemitic and non-antisemitic.
Through quantitative analysis, we identify paradoxical behaviors in some models that may help facilitate strategic interpretability audits to assess their robustness and reliability.
At last, we conduct a qualitative analysis to identify cases where all LLMs misclassify antisemitism. \textbf{Below, we summarize our main findings and contributions across eight models we study:}
\begin{itemize}[noitemsep, topsep=0pt]

  \item We present the first systematic evaluation of LLMs for antisemitism detection, demonstrating differences in utility (refusal rates, ambiguity, and repetitive generation) and performance traceable to model selection (\S~\ref{apx:take-aways}).
  
  \item Across nearly all models, our engineered Guided-CoT consistently outperforms Zero-Shot and Zero-Shot-CoT, regardless of decoding strategy, model size, or reasoning capability (\S~\ref{sec:evaluations}).
  Using Self-consistency, Guided-CoT improves positive-class F1-scores by at least 0.03 up to 0.13 compared to Zero-Shot-CoT and reduces refusal rates to nearly 0\%, thus enhancing model utility.

  \item Providing additional context (in our case, the IHRA definition with contemporary examples as policy instead of a short definition) does not necessarily improve model performance under Zero-Shot or Zero-Shot-CoT prompts, with some models experiencing even a decrease in performance (\S~\ref{sec:additional-context-check}).
  In such cases where there is a need to provide a policy in the prompt, Guided-CoT can help.

  \item We introduce metrics to quantify model explanations and find that Zero-Shot prompts result in homogeneous responses across models, yet individual models distinguish between antisemitic and non-antisemitic cases significantly (\S~\ref{sec:quantify-llm-explanations}).
  In contrast, CoT-based prompts, especially Guided-CoT, highlight differences in explanations across all models, while these differences between positive and negative classes are not significant for most models.

  \item Qualitative analysis reveals that LLMs struggle to understand contextual cues in writing patterns (\S~\ref{sec:qualit-analysis}).
  LLMs label posts as antisemitic solely because they contain stereotypical or offensive terms; additionally, LLMs mislabel quoted text and news-style reports, as well as neutral or critical opinions.
  Interestingly, LLMs flag typos (e.g.,`kikes' intended as `likes') and proper nouns (e.g., `Kiké') that resemble slurs as antisemitic.

\end{itemize}

\section{Background \& Related Work}

\subsection{Defining Antisemitism}

Previous work on antisemitism (and other forms of hate speech) has used a variety of definitions, which comes with a degree of subjectivity, with many being ad-hoc and directly developed by researchers or used only within academia.
This paper uses the definition of antisemitism created by the IHRA: \textit{``Antisemitism is a certain perception of Jews, which may be expressed as hatred toward Jews. Rhetorical and physical manifestations of antisemitism are directed toward Jewish or non-Jewish individuals and/or their property, toward Jewish community institutions and religious facilities.''}
Along with the definition, the IHRA also includes several contemporary examples of antisemitism (see Fig.~\ref{fig:ihra-definition}).
Although the IHRA's definition has been controversial, it has been widely adopted, including by the U.S. Department of State~\citep{ihra-usa}, numerous universities integrating it into discrimination policies~\citep{mosesYaleAddsContested2025, smithHarvardAgreesAdopt2025}, and legislative bills at the state level~\citep{hanshawCriticsRaiseFree} referencing it to address antisemitism.

\textbf{Potential issues with the definition.} We acknowledge that concerns about free speech have been raised regarding the IHRA's definition, primarily due to its explanatory note and contemporary examples, especially ``Denying the Jewish people their right to self-determination, e.g., by claiming that the existence of a State of Israel is a racist endeavor.''
At the same time, prior to the examples, the explanatory note says that  ``manifestations [of antisemitism] might include the targeting of the state of Israel, conceived as a Jewish collectivity'' but also that ``criticism of Israel similar to that leveled against any other country cannot be regarded as antisemitic.''
Some may see these statements as contradictory or argue that claiming Israel's existence is a racist endeavor is not antisemitic.
While we briefly discussed potential issues, understanding how LLMs can be used to understand this definition is critical due to their real-world applications.

\begin{figure*}[t]
\centering
\includegraphics[width=0.88\linewidth]{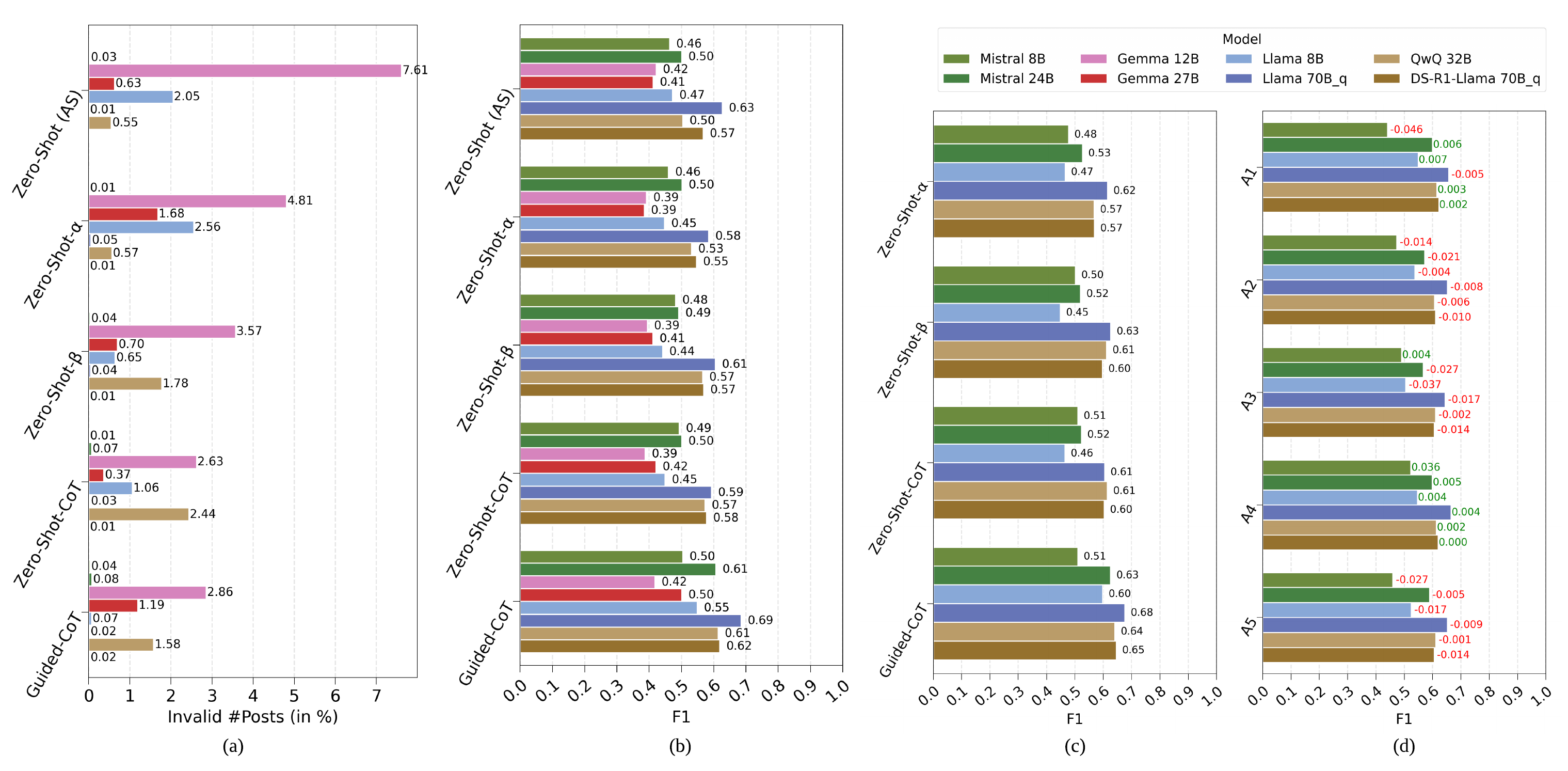}
\caption{
(a) Percentage of invalid responses per model by prompting technique.
(b) Greedy decoding.
(c) Self-consistency.
(d) Guided-CoT ablation.
Red indicates performance drop when excluding a thought from Guided-CoT; green indicates improvement.
\textbf{Metrics:} F1-score from positive class.
}
\label{fig:eval-all-in-one}
\end{figure*}

\subsection{Related Work}

Numerous datasets have been published to facilitate hate speech detection research~\citep{mathew2021hatexplain, lin2023toxicchat, hartvigsen2022toxigen, nghiem2024hatecot}, and researchers have studied it extensively across many social media platforms~\citep{jahan2024comprehensive, casula2024target, tahmasbiGoEatBat2021, zannettou2020quantitative, antypas2023robust}.
However, training and deploying adaptable classifiers to the evolving social media landscape remains challenging.

Although LLMs' capabilities in detecting harmful content have been studied~\citep{kumarage2024harnessing, guo2023investigation, kikkisetti2024using, nirmal2024towards}, effective moderation requires an understanding of socio-cultural context~\citep{kumar2024socio}, nuanced definitions and policies~\citep{huang2025content, goldberg2024ai} related to hate speech.
Hate-speech moderation faces fairness and bias-related issues: early work shows demographic bias in classifiers \citep{sap2019risk}, and LLM-based systems yield inconsistent judgments across models \citep{fasching2025model}.
LLMs can be oversensitive and poorly calibrated for implicit hate \citep{zhang2024don}, and remain vulnerable to adversarial prompting that elicits toxic outputs despite guardrails \citep{dutta2024down}.
~\citet{aldreabi2023enhancing} showed that changes in the definition lead to variations in toxicity scores.
In LLM-related benchmarking study, ~\citet{balachandran2024eureka} examined model discrepancies in toxicity detection, reporting that almost all models they evaluated show an accuracy gap of 10\% for detecting toxicity targeting Jewish individuals.
Despite the wide adoption of IHRA's definition by institutions globally, to our knowledge, this study explores the first-ever LLMs' capabilities in detecting antisemitism using IHRA's definition as a policy.

Researchers have improved LLM performance on downstream tasks through CoTs~\citep{wei2022chain, kojima2022large, vishwamitra2024moderating}, Self-consistency~\citep{wang2022self, wan2024reasoning}, emotion prompts~\citep{li2023emotionprompt}, and echo prompts~\citep{mekala2023echoprompt}.
~\citet{kolla2024llm, goyal2025momoe} and~\citet{zheng2024hatemoderate} used Reddit and Facebook community guidelines, respectively, highlighting the strengths and limitations of LLMs in applying comprehensive policy definitions.
Furthermore, ~\citet{balachandran2024eureka} observed that while some models achieve high accuracy in toxicity detection, they also exhibit high refusal rates, limiting their practical utility.

Explainability remains a central aspect in utilizing LLMs.
~\citet{yang2023hare} shows that step-by-step LLM rationales can improve implicit hate detection; however, ~\citet{di2024explanation} finds that LLM-generated explanations often lack usefulness and trustworthiness, warranting caution.
~\citet{shaikh2022second} shows that Zero-Shot-CoT can increase the production of harmful or undesirable output in socially sensitive tasks, but can also decrease it with improved instructions.
To explore this further, our study introduces Guided-CoT, a CoT-like prompting designed to enhance the understanding of in-context policy, performance, and utility of sensitive classification tasks.
Additionally, we propose quantitative metrics for analyzing LLM-generated rationales for better interpretability and reliability for content moderation.

\section{Experimental Setup}

\textbf{LLMs\footnote{NB: Not all models responded to classification; some refused to classify, while others provided labels other than Yes or No.
We exclude the posts with invalid responses to ensure direct comparison between models for individual analysis.
For simplicity, we refer to models by name and size (e.g., Gemma 12B), DeepSeek-R1-Distill-Llama as DS-R1-Llama, Zero-Shot as ZS, and use `\_q' to denote quantized versions.}.} We evaluate eight open-source LLMs from five families, including reasoning and non-reasoning: Gemma-3 12B, Gemma3 27B, Mistral-2410 8B, Mistral-2501 24B, Llama-3.1 8B, Llama-3.1 70B, QwQ 32B, and DeepSeek-R1-Distill-Llama 70B (see Table~\ref{apx:model-attributes}).
We use quantized versions for larger models (i.e., Llama-3.1 70B and DeepSeek-R1-Distill-Llama 70B).

\textbf{Prompting Techniques.} \citet{tam2024let} finds that model performance significantly declines when restricted to generating content under formatting constraints.
Therefore, we design our prompts that allow models to generate free-form text to start the response, followed by a summary and a classification label at the end of its response.
Overall, we explore Zero-Shot and CoT~\citep{schulhoff2024prompt} prompting techniques and propose a new prompt design, Guided-CoT.

\textbf{ZS-(AS):} We create a ZS-(AS) to evaluate the model's foundational knowledge in detecting antisemitic content and compare it to more sophisticated prompting techniques, treating this as a baseline (see Fig.~\ref{fig:baseline-prompt}).

\textbf{Zero-Shot (IHRA):} We instruct models to classify a social media post by designing two Zero-Shot variants:

\textbf{1) ZS-{\LARGE $\alpha$}:} Model classifies a post as ``Yes'' (antisemitic) or ``No'' (not antisemitic) according to the IHRA's definition without any additional context, like contemporary examples (see Fig.~\ref{fig:zeroshot-alpha-prompt}).

\textbf{2) ZS-{\LARGE $\beta$}:} Building on the ZS-$\alpha$, this includes IHRA's definition with contemporary examples (767 tokens) in the system prompt, with the same user prompt as ZS-$\alpha$ (see Fig.~\ref{fig:zeroshot-beta-prompt}).

\textbf{ZS-CoT:} This prompting technique uses the phrase ``\textit{Let's think step by step},'' encouraging the model to reason through the classification process.

\textbf{Guided-CoT (Ours):} We design this CoT-like prompt to mimic human annotators' thought processes (e.g., develop a codebook to guide the annotation process~\citep{halterman2024codebook, ling2022slapping}) in labeling posts for antisemitism.
We guide the model through the steps (see Fig.~\ref{fig:guided-cot-prompt}): 
1) Break down the post to understand the author's written pattern, 
2) Identify hidden tones like sarcasm, criticism, coded language, ambiguous phrasing, quoted statements, news, and reporting, 
3) Check if the author is calling out to promote Antisemitism or spreading awareness, 
4) Evaluate whether the post aligns with the IHRA's definition and contemporary examples, 
5) Based on the above steps, analyze whether the author's stance toward the Jewish community is favorable.

\textbf{Dataset \& Metrics.} We use an existing dataset~\citep{jikeli2023antisemitic, jikeli2024annotating} labeled by human annotators based on the IHRA definition.
This dataset consists social media posts collected from Twitter (now known as X) covering diverse conversations about Jews, Israel, and antisemitism between January 2019 and April 2023.
It contains 11,315 social media posts (i.e., tweets), of which 9,362 are non-antisemitic, and 1,953 (17\%) are antisemitic.
As the dataset highly unbalanced, we select F1, precision, and recall for the positive class to measure the model performance.

\section{Evaluations}
\label{sec:evaluations}

\subsection{Differences in Generated Responses}
\label{sec:diff-in-responses-main}

We first examine differences in LLMs' responses based on greedy decoding (temperature set to zero), focusing on cases where models refuse or fail to provide valid responses.
Based on our prompt design, response contain three things: free-form text, a summary, and a classification label.

We categorize model responses into three groups:
1) \textbf{Valid} responses include a summary and an explicit antisemitic label (Yes or No), 
2) \textbf{Indeterminate} responses include a summary and a decision label, but the label is neither Yes nor No. E.g., ``Potentially,'' ``Ambiguous,'' or ``Cannot be determined without context,'' and 
3) \textbf{Failure} responses either exceed the maximum token length (set to 2048) or the model explicitly refuses to respond, indicated failure-exceed and failure-refusal, respectively.

Four models, Gemma 12B (15.34\% posts), Gemma 27B (3.55\%), Llama 8B (3.92\%), and QwQ 32B (5\%), frequently generate invalid responses (either type of Indeterminate or Failure); percentages are calculated across all five prompting variants (see Fig.~\ref{fig:eval-all-in-one} (a)).
Prompting style substantially affects these behaviors: Guided-CoT reduces Llama 8B's refusal rate to only 0.07\% compared to Zero-Shot and ZS-CoT, while QwQ 32B exceeds the 2048-token limit primarily when the IHRA definition with contemporary examples is added.
In Apx.~\ref{apx:diff-in-generated-responses}, we discuss differences across other axes.

\subsection{Deterministic Evaluation}
\label{sec:deterministic-eval}

Some models generate invalid responses, so we exclude those posts and responses from the analysis.
We evaluate models with greedy decoding for valid responses (see Fig.~\ref{fig:eval-all-in-one} (b)), a subset of 8,555 posts (7,031 non-antisemitic and 1,524 antisemitic).

We examine changes in model's performance from ZS-(AS) to ZS-$\alpha$, from ZS-$\alpha$ to ZS-$\beta$, and so on.
ZS-(AS) baseline reflects model's inherent understanding of antisemitism, as we do not provide an explicit definition.
Under ZS-(AS), Llama 70B\_q performs best, while Gemma models perform worst (high recall with low precision), indicating a bias in labeling content as antisemitic (see Fig.~\ref{fig:common-points-models-performance-greedy}).
Adding the IHRA definition (ZS-$\alpha$) yields mixed performance, most models either slightly decrease or remain unchanged, except QwQ 32B's performance increases ($+0.03$).
The IHRA definition, including contemporary examples (ZS-$\beta$), improves performance for most models, except Llama 8B and Mistral 24B.
To investigate this behavior, we prompt all models to define antisemitism and find that only Llama 8B, Mistral 24B, and DS-R1-Llama 70B\_q do not reference the IHRA definition, potentially resulting in less exposure during training (details in Apx.~\ref{apx:antisemitism-definition}).
ZS-CoT yields mixed performance, except Llama 70B\_q ($-0.02$).
Guided-CoT improves performance across all models.
Overall, most models gain performance from more straightforward to sophisticated prompts.

NB: We exclude Gemma models for low performance and potential bias in labeling, and also exclude the ZS-(AS) baseline from further analysis.

\subsection{Non-deterministic Evaluation}
\label{sec:non-det-eval}

Previous studies indicate that prompt design~\citep{sclar2023quantifying, atreja2024prompt}, non-determinism~\citep{song2024good}, and hyperparameters~\citep{renze2024effect} can affect LLM performance.
While we engineered our prompts by experimenting, we can leverage the Self-consistency method~\citep{wang2022self}, which has shown robustness to imperfect prompts and can help us compare a model's true performance over a complete dataset.

We run 30 inferences on all 11,315 data points (temperature 0.6, top\_p 0.9), determining final classification decisions by majority voting.
Llama 70B\_q achieves the highest overall F1-score (0.66), with an increase of 0.09 compared to ZS-{$\beta$} and ZS-CoT (see Figs.~\ref{fig:eval-all-in-one} (c) \&~\ref{fig:model-performance-SC}).
Interestingly, the mid-sized Mistral 24B performs similar to the reasoning model QwQ 32B ($F1=0.58$).
Guided-CoT improves performance across all models compared to ZS-CoT, with a substantial 0.13 increase for Llama 8B.
Our results confirm that Guided-CoT paired with self-consistency consistently yields higher performance, regardless of model size or reasoning capability.

\textbf{Comparison to Existing Work.} Our work uses Version 3~\citep{ihra-dataset-v3} of ~\citet{jikeli2023antisemitic}'s dataset.
\citet{becker2024antisemitism} used Version 1~\citep{ihra-dataset-v1} of the same dataset, comprising 6,941 posts, to evaluate their fine-tuned GPT-3.5 model trained to detect antisemitism.
We compare our best-performing model on Version 1 to the evaluation results from ~\citet{becker2024antisemitism}'s study (see Table~\ref{tab:all-guided-cot-stats}).
Using Guided-CoT, Llama 70B\_q achieves an F1 Score of 0.72 with greedy decoding and 0.73 with self-consistency, outperforming fine-tuned GPT-3.5 (F1=0.70).
NB: We do not have access to ~\citet{becker2024antisemitism}'s training dataset or the fine-tuned GPT-3.5; we rely on the metrics reported in their study for comparison.

\subsection{Guided-CoT Ablation}
\label{sec:guided-cot-ablation}

We now conduct an ablation study for Guided-CoT to identify which thoughts contribute the most to models' performance, excluding one thought at a time (out of 5 thoughts).
For example, prompt ``A1'' excludes thought A1 while retaining thoughts A2, A3, A4, and A5 (more details in Apx.~\ref{apx:guided-cot-more}).
For ablation analysis, we have a subset of 10,442 posts (8,594 non-antisemitic and 1,848 antisemitic), excluding the posts with invalid responses in one or more ablation conditions.
To measure performance, we calculate the difference (delta) over F1-score, e.g. A1 = \{A2, A3, A4, A5\} Vs. Guided-CoT = \{A1, A2, A3, A4, A5\}.
For instance, Mistral 8B scores 0.44 with prompt A1 and 0.49 with Guided-CoT, showing about 0.05 decrease in performance when A1 is excluded (see Fig.~\ref{fig:eval-all-in-one} (d)).

For all models, important thoughts to improving performance are: 
1) explicitly mention checking for sarcasm and criticism (A2),
2) explicitly instruct to check whether a post promotes antisemitism or spreads awareness (A3), and
3) a thought that encourages the model to reflect on analyzing whether the author's overall stance is favorable toward the Jewish community (A5).

Conversely, breaking posts into chunks (A1) does not improve performance, except for Mistral 8B, which benefits substantially from it (0.05 increase in F1).
Surprisingly, instructing explicit alignment with the IHRA definition (A4) may have improved performance for most models if excluded from Guided-CoT.

\subsection{Impact of Additional Context across Decoding and Prompting}
\label{sec:additional-context-check}

We previously discussed performance changes resulting in the progression from simpler to more sophisticated prompts.
Now, we focus on how models handle additional context; the key research question is: \textbf{\textit{How does performance change when using just the IHRA definition versus adding a policy (i.e., the IHRA definition with contemporary examples) across different configurations?}}

\textbf{Setup:} We compare three configurations: 1) Greedy, 2) Sampling-based, and 3) Self-consistency.
We use the same hyperparameters for Greedy decoding and Self-consistency as described earlier.
For the sampling setup, we run the dataset 5 times (temperature of 0.6 and top\_p of 0.9) and average the results.
For this analysis, we include 8,624 of the 11,315 data points (7,086 non-antisemitic and 1,538 antisemitic), excluding the posts with invalid responses.

\begin{figure}[t]
\centering
\includegraphics[width=0.80\linewidth]{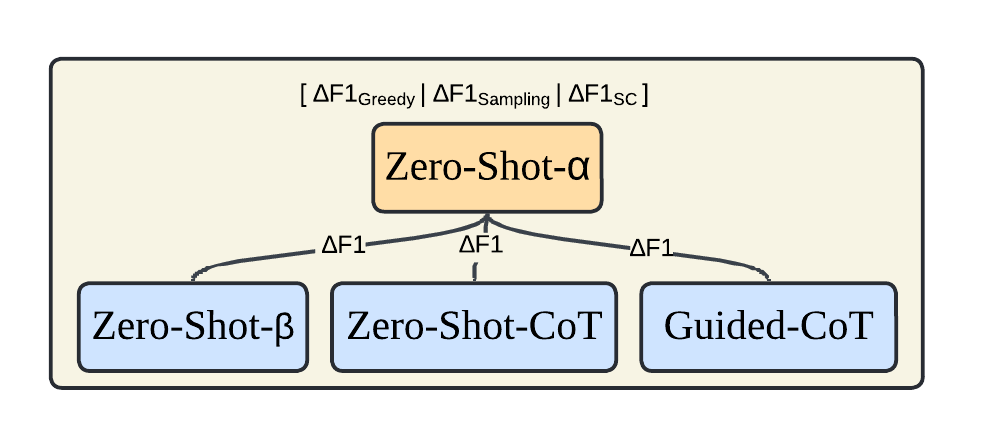}
\caption{Comparing a transition from ZS-{\boldmath $\alpha$} to other prompting across different decoding strategies.}
\label{fig:compare-decode-to-prompt}
\end{figure}

We use ZS-{$\alpha$} as the base and pair it with other prompting to observe performance differences ({$\Delta$}), calculated by subtracting the F1 score from ZS-$\alpha$ to a selected prompting (see Fig.~\ref{fig:compare-decode-to-prompt}).
For example, the ZS-{\boldmath $\alpha$} to ZS-{\boldmath $\beta$} pair looks at the performance change when we add additional context (the IHRA definition with contemporary examples).

\begin{figure}[t]
\centering
\includegraphics[width=0.98\linewidth]{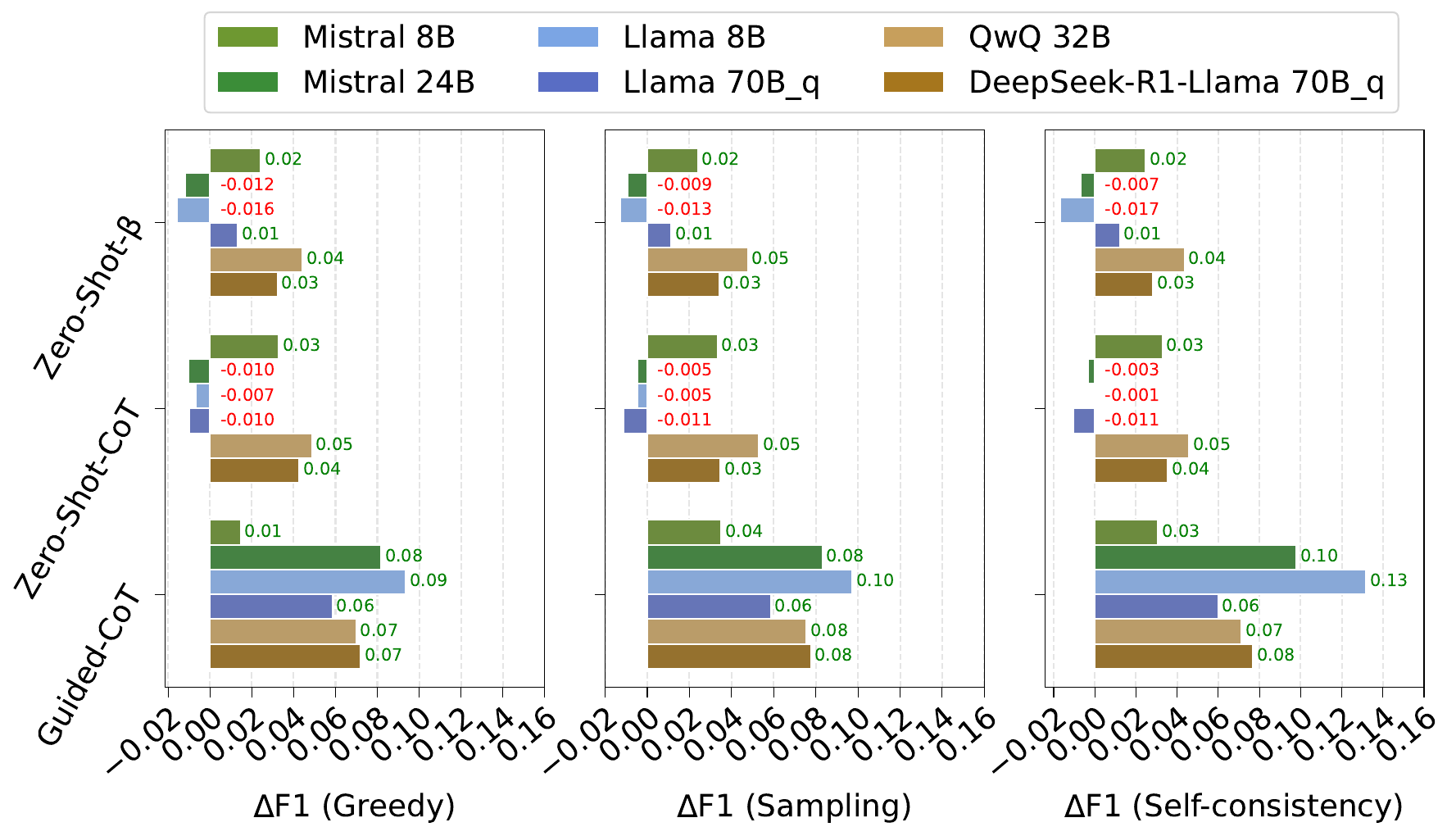}
\caption{A {$-\Delta$} indicates a decrease in performance, and a {$+\Delta$} denotes an increase when transitioning from ZS-{\boldmath $\alpha$} to the selected prompting.}
\label{fig:compare-decode-to-prompt-results}
\end{figure}

The performance differences across prompting techniques remain consistent regardless of the decoding strategy (see Fig.~\ref{fig:compare-decode-to-prompt-results}).
For instance, transitioning from ZS-{$\alpha$} to ZS-{$\beta$} decreases Llama 8B and Mistral 24B performance across all three configurations.
Similarly, ZS-{$\alpha$} to ZS-CoT does not improve performance for three models: Llama 8B, Mistral 24B, and Llama 70B\_q.
ZS-{$\alpha$} to Guided-CoT consistently improves performance for all models and configurations.
Overall, all models gains the performance with Guided-CoT, except for Mistral 8B with greedy decoding, where the improvement is not as good as ZS-CoT.

\begin{figure}[t]
\centering
\includegraphics[width=0.98\linewidth]{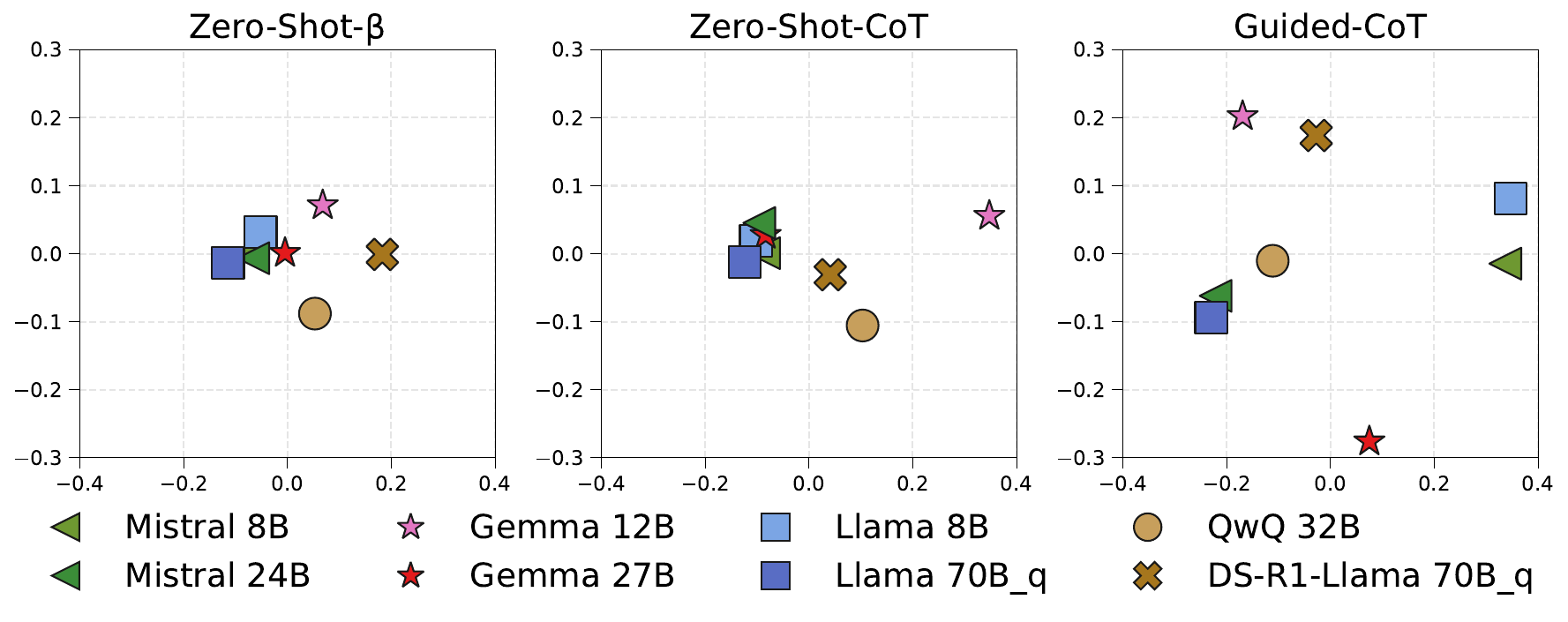}
\caption{Projection of each model's Semantic Distance Vector (SDV) into 2D space using PCA. X and Y axes represent the first two principal components.}
\label{fig:pca-clustering}
\end{figure}

\section{Quantifying Models' Explanations}
\label{sec:quantify-llm-explanations}

We examine the differences in the model-generated responses; the key research questions are: \textbf{\textit{To what extent do LLMs generate distinct explanations from other models? How do these explanations differ when they label a post as antisemitic versus non-antisemitic?}}
Understanding these differences is important for model selection, as their ability to explain can help us identify biases or assess reliability.

\textbf{Setup:} This analysis includes 9,725 posts with valid responses from eight models under three prompting strategies (ZS-$\beta$, ZS-CoT, Guided-CoT) using greedy decoding.
We generate embeddings of LLM-generated responses using \texttt{llm2vec}~\citep{behnamghader2024llm2vec} and reduce them to 15 dimensions via UMAP~\citep{mcinnes2018umap}.
NB: We exclude intermediate ``thinking'' tokens for reasoning models to embed responses.

\subsection{Cross-Model Difference Analysis}

The differences examined through this analysis highlight the organic variations in generated responses, capturing rationale, a classification decision, and possible ubiquitous templating patterns, especially when using Guided-CoT.
We recognize that simply embedding a complete response as it is may not accurately reflect contextual differences, as responses may follow a designed format and may include elements from the original social media posts.
However, this approach provides a useful first view on how models diverge.

\vspace{-0.5\baselineskip}
\begin{align}
  d^{A,n} &= \mathcal{D}_c(\mathbf{e}_p^A,\mathbf{e}_q^n) \label{eq:cos_dist}\\
  SDV_A   &= \bigl[\mathrm{median}(d^{A,n})\bigr]_{n\in\mathcal{N}} \label{eq:median_vector}
\end{align}

For each model pair (e.g., Llama 8B and Mistral 8B), we compute the normalized cosine distance (0: highly similar to 1: substantially different) between their embedded responses.
At this step, we have a distribution $d^{A,n}$, shown in Eq.~\ref{eq:cos_dist}, where $\mathcal{N}$ is the set of the other seven models ($n \in \mathcal{N}$), $p \in \{\text{responses by model }A\}$, $q \in \{\text{responses by model }n\}$, ${e}_p^A$ represents embeddings of all responses generated by model $A$, and $\mathcal{D}_c$ is a cosine distance (see Fig.~\ref{fig:zs-beta-facet}, Fig.~\ref{fig:zs-cot-facet}, Fig.~\ref{fig:guided-cot-facet}).
Now, we represent each model by a 7-dimensional vector, which we call \textbf{Semantic Distance Vector (SDV)}. 
For instance, $SDV_A$ indicates model A’s semantic median-distance to every other model (see Eq.~\ref{eq:median_vector}).

\vspace{-0.5\baselineskip}
\begin{align}
  SCMD_{rank(A)} &= \frac{1}{|\mathcal{N}|}\sum SDV_A \label{eq:average_median_rank}\\
  SCMD_{cluster} &= \mathrm{PCA}\!\left(\bigl[\mathrm{SDV}_n\bigr]_{n\in\mathcal{N}}\right) \label{eq:average_median_cluster}
\end{align}

Finally, we calculate \textbf{Semantic Cross-Model Divergence (SCMD)}, using \textbf{SDV}, capturing the model's overall divergence across other models. Now, we examine differences along two axes:

\textbf{1) Rank: which model is most or least diverged from other models?}

We use $SCMD$ to rank models (see Eq.~\ref{eq:average_median_rank}).
A high value indicates a distinctive explanatory pattern that diverges from the norm, while a low value reflects alignment with other models' responses.
For instance, SCMD of Mistral 8B is 0.12 (See Table~\ref{tab:sorted-model-distance}).
CoT-like prompting has a higher SCMD for any given model than Zero-Shot.
However, no consistent ranking pattern emerges between models.
Llama 70B\_q consistently demonstrates the lowest SCMD, indicating its explanations closely align with other models.

\textbf{2) Cluster: which models are closer to or further apart from each other in their explanations?}

Using Principal component analysis (PCA), we project each model's 7-dimensional vector, $SDV$ (Eq.~\ref{eq:average_median_cluster}), into 2D space, assuming that models that share explanatory patterns may appear closer regardless of the differences in predictions (see Fig.~\ref{fig:pca-clustering}).
Under ZS-{$\beta$}, models' responses appear relatively closer, except for both reasoning models and Gemma 12B.
ZS-CoT further separates both reasoning models and Gemma 12B from the rest.
Guided-CoT further amplifies divergence for all models except Llama 70B\_q, and Mistral 24B appears closer.

Overall, Zero-Shot prompts yield homogeneous responses and Guided-CoT surfaces latent stylistic differences.
Semantic divergence offers insights beyond accuracy, informing model selection where explanatory power is important for nuanced tasks like hate-speech detection.

\begin{figure}[t]
  \centering
  \begin{subfigure}[t]{\linewidth}
    \centering
    \includegraphics[width=0.98\linewidth]{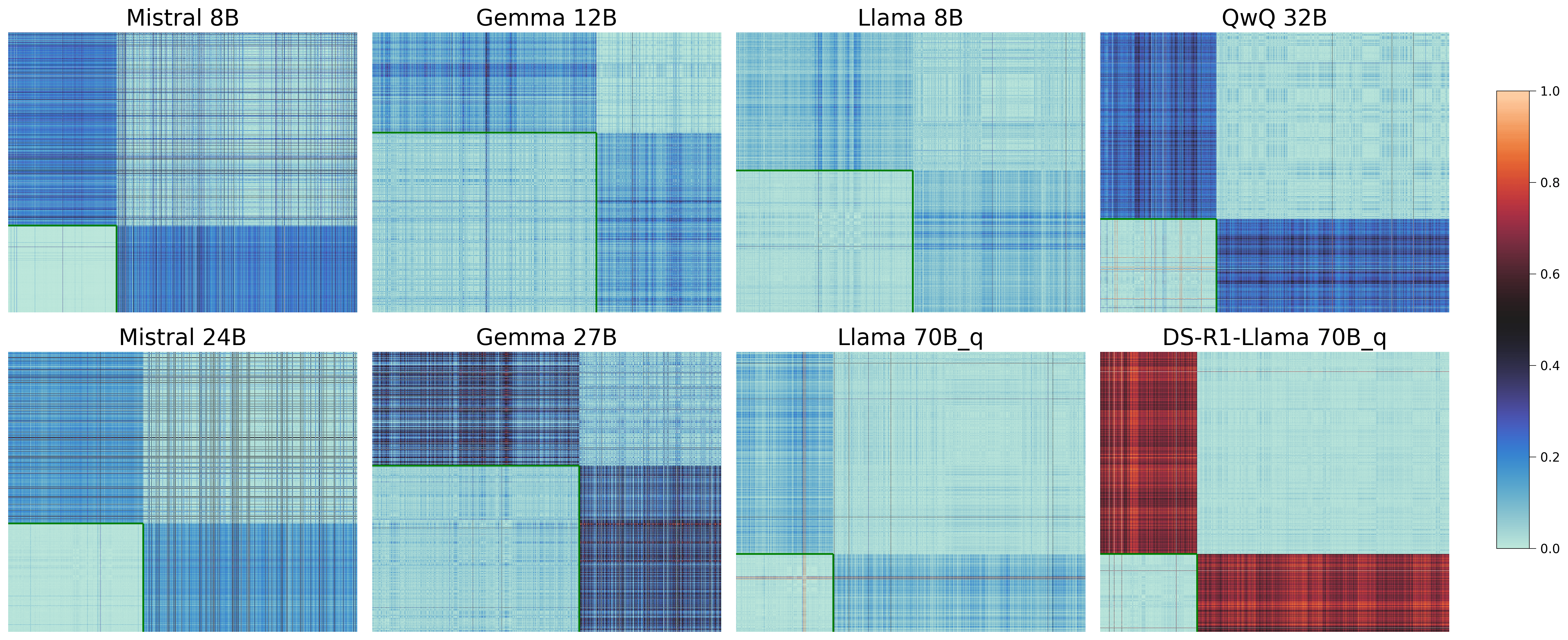}
    \caption{ZS-$\beta$}
    \label{fig:zs-beta-heatmap}
  \end{subfigure}

  \begin{subfigure}[t]{\linewidth}
    \centering
    \includegraphics[width=0.98\linewidth]{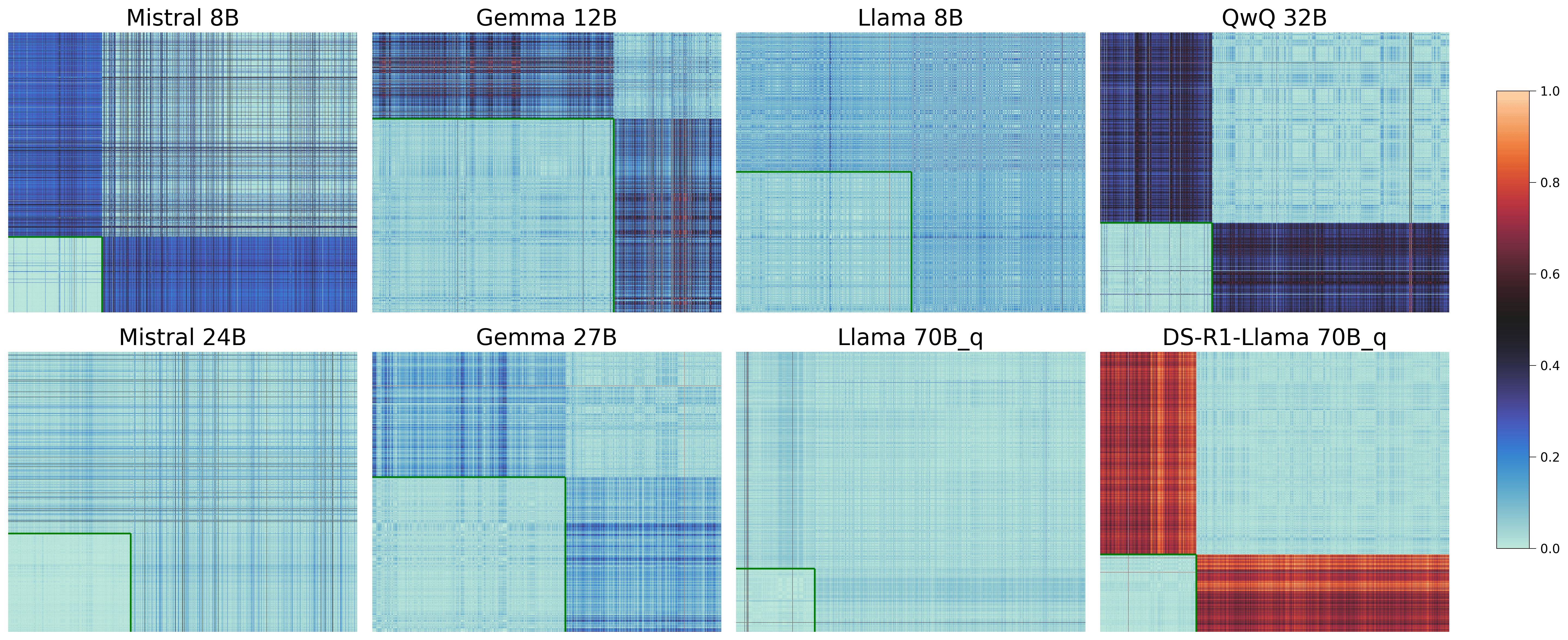}
    \caption{ZS-CoT}
    \label{fig:zs-cot-heatmap}
  \end{subfigure}

  \begin{subfigure}[t]{\linewidth}
    \centering
    \includegraphics[width=0.98\linewidth]{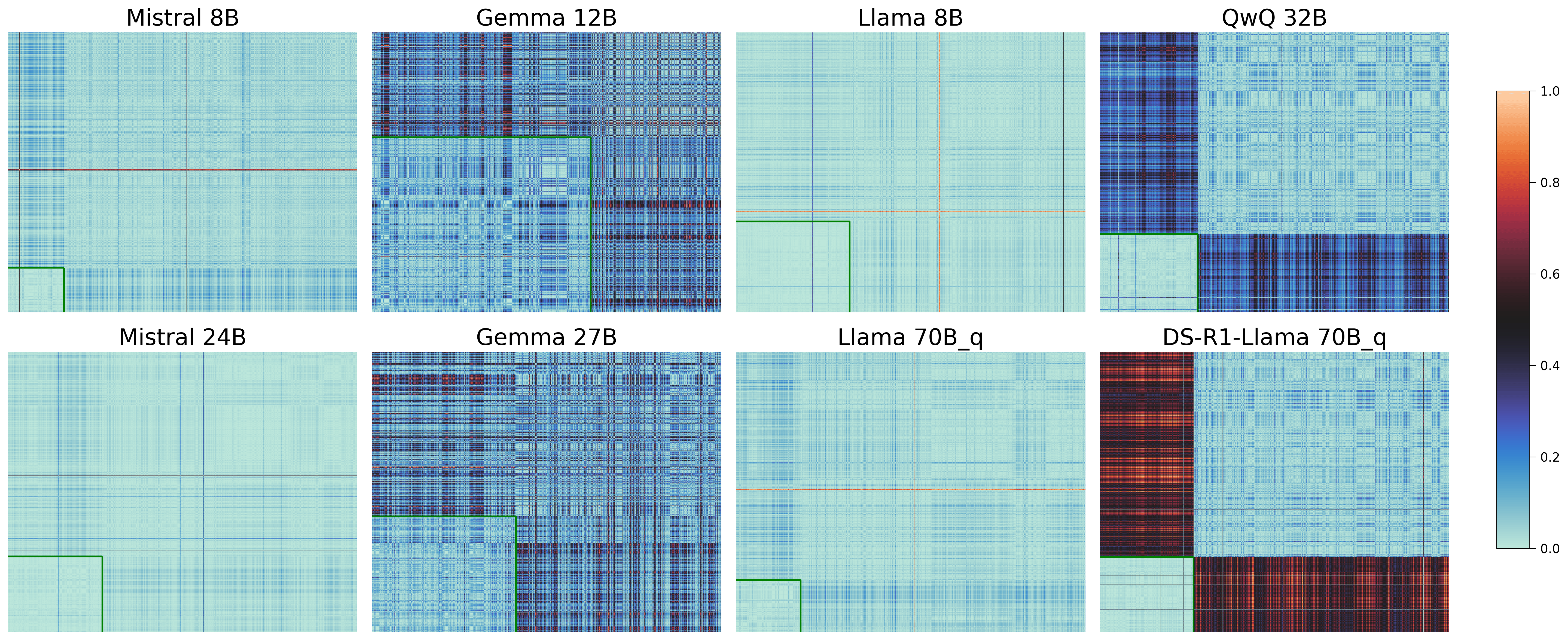}
    \caption{Guided-CoT}
    \label{fig:guided-cot-heatmap}
  \end{subfigure}
  \caption{Heatmap of model's responses grouped by antisemitic prediction (bottom-left) vs. non-antisemitic (top-right).}
  \label{fig:heatmap-comparison}
\end{figure}

\begin{table*}[t]
\tiny
\centering
\begin{tabular}{l l|l|l l|l|l l|l|l}
\toprule
 & \multicolumn{3}{c}{$H_0$: $D^{+}$ = $D^{-}$} 
 & \multicolumn{3}{c}{$H_0$: $D^{+}$ <= $D^{-}$, $H_a$: $D^{+}$ > $D^{-}$} 
 & \multicolumn{3}{c}{$H_0$: $D^{+}$ >= $D^{-}$, $H_a$: $D^{+}$ < $D^{-}$} \\
\cmidrule(lr){2-4}\cmidrule(lr){5-7}\cmidrule(lr){8-10}
\textbf{Model} & 
  \textbf{ZS-{$\beta$}} & \textbf{ZS-CoT} & \textbf{G-CoT} & 
  \textbf{ZS-{$\beta$}} & \textbf{ZS-CoT} & \textbf{G-CoT} & 
  \textbf{ZS-{$\beta$}} & \textbf{ZS-CoT} & \textbf{G-CoT} \\
\midrule

Mistral 8B
  & 0.11$^{***}$ & 0.11$^{***}$ & 0.00
  & 0.12$^{***}$ & 0.00 & -
  & 0.55$^{***}$ & 0.56$^{***}$ & - \\

Mistral 24B
  & 0.03$^*$ & 0.01 & 0.03$^*$
  & 0.06$^*$ & - & \cellcolor{red!30}0.06$^{**}$
  & 0.86$^{***}$ & - & \cellcolor{red!30}0.86$^{***}$ \\

Gemma 12B
  & 0.43$^{***}$ & 0.44$^{***}$ & 0.12$^{***}$
  & 0.36$^{***}$ & \cellcolor{red!30}0.36$^{***}$ & 0.10$^{***}$
  & 0.51$^{***}$ & \cellcolor{red!30}0.51$^{***}$ & 0.00 \\

Gemma 27B
  & 0.05$^{***}$ & 0.07$^{**}$ & 0.00
  & 0.06$^{**}$ & 0.01 & -
  & 0.19$^{***}$ & 0.19$^{***}$ & - \\

Llama 8B
  & 0.06$^{***}$ & 0.11$^{***}$ & 0.00
  & 0.11$^{***}$ & \cellcolor{red!30}0.11$^{***}$ & -
  & 0.07$^{**}$ & \cellcolor{red!30}0.07$^{***}$ & - \\

Llama 70B\_q
  & \cellcolor{green!30}0.02 & \cellcolor{green!30}0.04 & \cellcolor{green!30}0.01
  & - & - & -
  & - & - & - \\

QwQ 32B
  & 0.04$^{**}$ & 0.04 & 0.00
  & 0.06$^*$ & - & -
  & 0.95$^{***}$ & - & - \\

DS-R1-Llama 70B\_q
  & 0.07$^{***}$ & 0.09$^{***}$ & 0.07$^{***}$
  & 0.26$^{***}$ & 0.00 & 0.26$^{***}$
  & 0.95$^{***}$ & 0.95$^{***}$ & 0.01 \\

\bottomrule
\end{tabular}
\caption{A difference is considered significant only if indicated by an asterisk. The numbers are KS statistics.
$P-value$ < 0.001, 0.01, and 0.05 are marked with ***, **, and *, respectively.}
\label{}
\end{table*}

\subsection{Intra-Model Difference Analysis}

We now investigate intra-model differences, specifically how a single model's responses differ when labeling posts as antisemitic versus non-antisemitic.
This analysis isolates differences within models rather than between them.

\textbf{Differences}: For each model, we compute a pairwise cosine distances (normalized [0,1]) from 15-dimensional embeddings, producing an $N \times N$ distance matrix ($N=9{,}725$), reordered by predicted label, grouping ``antisemitic'' and ``non-antisemitic'' responses.
If a model's explanations are semantically similar, the heatmap shows cooler colors; if diverse, it shows warmer colors (see Fig.~\ref{fig:heatmap-comparison}).
We visualize three groups ($\mathcal{G}$) for discussion:
\textbf{1) ${\boldmath \mathcal{G}^{++}}$:} LLM-generated responses of all posts classified as antisemitic,
\textbf{2) $\mathcal{G}^{--}$:} Responses of all posts classified as non-antisemitic, and
\textbf{3) $\mathcal{G}^{+-}$:} Responses of all posts classified as antisemitic compared to classified as non-antisemitic.

Under ZS-$\beta$, we observe clear distinctions in $\mathcal{G}^{+-}$, especially in reasoning models.
In contrast, $\mathcal{G}^{++}$ and $\mathcal{G}^{--}$ appear visually similar, except for the Mistral family.
For the Mistral family, explanations differ visually between antisemitic and non-antisemitic classifications but remain consistent within each group.
Under ZS-CoT, the separation in $\mathcal{G}^{+-}$ weakens for the Llama models and Mistral 24B, and explanations in $\mathcal{G}^{++}$ and $\mathcal{G}^{--}$ become highly similar, with $\mathcal{G}^{++}$ showing higher similarity than $\mathcal{G}^{--}$.

With Guided-CoT, we observe interesting patterns:
1) The Mistral and Llama families show uniform behavior within groups.
We interpret this positively, as Guided-CoT encourages adhering to the policy-oriented IHRA guideline, regardless of the classification accuracy.
2) For the Gemma family, exhibiting distinct, tile-like patterns for $\mathcal{G}^{--}$ and $\mathcal{G}^{++}$, with $\mathcal{G}^{++}$ being more similar than $\mathcal{G}^{--}$.
We hypothesize that the Gemma models may differentiate responses based on different topics in input content, but further qualitative assessment is needed.
3) Reasoning models display clear separation across all groups in their heatmaps.

\textbf{Significance}: We conduct a significance test to determine if the differences observed through heatmap are statistically significant (more details in Apx.~\ref{apx:significance-details}).
Because the $\mathcal{G}^{++}$ and $\mathcal{G}^{--}$ groups differ in size, we randomly sample (k=1,500) responses per group without replacement.
Next, we calculate the cosine distance between every pair of responses within each group.
From these distances, we form two distributions: $D^{+}$ and $D^{-}$.
For instance, the $D^{+}$ represents the average distances between responses labeled as positive.

Under the Zero-Shot setting, all models except Llama 70B\_q show significant differences, indicating that $\mathcal{G}^{++}$ differ from those for $\mathcal{G}^{--}$.
With ZS-CoT, the differences are significant for only four models, and with Guided-CoT for only three.
Guided-CoT neutralizes response differences in models like Mistral 8B, Gemma 27B, and Llama 8B, compared to ZS-CoT, operating as a stylistic regularizer.
This behavior is meaningful as it may reduce hallucinations and biases, allowing models to focus on the task at hand.
Interestingly, Llama 70B\_q shows no significant differences between explanations for $\mathcal{G}^{++}$ and $\mathcal{G}^{--}$ across prompting.

Additionally, we use a one-sided KS-test to compare $D^{+}$ and $D^{-}$, checking if semantic distances for $\mathcal{G}^{++}$ are significantly larger or smaller than $\mathcal{G}^{--}$.
NB: We perform this test only where $D^{+}$ and $D^{-}$ are significantly different.
Under ZS-$\beta$, all models reject one-sided KS-test, indicating that $D^{+}$ is not larger or smaller than $D^{-}$.
This paradoxical behavior suggests bimodal or skewed distributions (crossing CDFs) in semantic differences, furthering the need for qualitative analysis to reveal potential biases or failure cases.
Under ZS-CoT, this crossing behavior is observed only for Gemma 12B and Llama 8B, and under Guided-CoT, only for Mistral 24B.
These three models require focused scrutiny, as their antisemitic responses cannot simply be categorized as uniformly more or less cohesive; rather, they exhibit distinct subgroupings, extremely cohesive for some responses and highly divergent for others.

\section{Misclassification Analysis}
\label{sec:qualit-analysis}

In this section, we qualitatively analyze the misclassified posts by sampling 260 posts (out of 8,555, for which we report performance in \S~\ref{sec:deterministic-eval}) that have the same predictions from all eight models but do not match the ground truth.
Two authors categorize these posts manually into: 1) Quotes/News-Reporting (Q/NR), 2) Tone/Language/Stereotypes, 3) Opinion, 4) Religious, 5) Sarcasm, and 6) Contextual Mismatch.
After both annotators independently completed the annotations (inter-rater agreement of Cohen's Kappa $0.75$ - considered high), we resolved the disagreements to finalize the categories.
\textbf{As a result, we find the category-wise distribution as:} Tone/Language/Stereotype (28\%); Q/NR (25\%); Opinion (21\%); Contextual Mismatch (8\%); Sarcasm (6\%); and Religious (6\%).
We detail the codebook in the Apx.~\ref{apx:codebook}.

We observe that 28\% of misclassified posts fall under Tone/Language/Stereotype, followed by 25\% under Q/NR.
For example, the post ``\textit{Weapons made in Israel are used for genocide, for ethnic cleansing L: https://t.co/NkCuaP3nkW C: https://t.co/2uyaexvQzw}'' criticizes actions of Israel; all models flag it as positive, yet it is not antisemitic because it does not invoke an antisemitic trope according to IHRA's definition.
Under Q/NR, the post ``\textit{Palestinian journalist Laila Ouda has been fired by France 24 for supporting her people's cause and exposing Israel's crimes and violations of international law. https://t.co/wsXvjK2asZ}'' illustrates a false negative.
Most false positives fall under Tone/Language/Stereotype, whereas most false negatives fall under Q/NR.

We also observe that user opinions contribute about 21\% of errors.
For instance, ``\textit{@CathyHigginsWY @VP @POTUS Whoaaaaaaaaaa. This makes me question my whole belief system. Next you're gonna tell me that Soros doesn't control all Jews and that rural folk aren't subsidized by the devil worshipping city fags who eat curry and smoke jazz cigarettes}'' is a false positive.
We identify a contextual mismatch type in 8\% of posts.
Typos where users write `kikes' instead of `likes' lead models to misclassify content as antisemitic (e.g., ``\textit{RT @Purbita9: \#AdiZaMountThisAward Aditi Rathore is daring she directly kikes adiza VMS}'').
Proper nouns containing `Kike' are also misinterpreted as a slur `kikes' (e.g., ``\textit{So sad I missed kikes booblehead night because of work but at least I got to see him walk off}'', a reference to the Dodgers bobblehead night for player Kiké Hernández).

Overall, models frequently struggle to assign a correct label when posts include offensive language, tone, or stereotypes, as well as quoted statements or opinions, posing a critical challenge to using these models in an automated fashion for social media content moderation.
Our findings complement the previous study~\citep{roy2023probing}.
Although experimenting with additional context is outside the scope of this work, prior studies~\citep{kumar2024watch} find that adding additional context improves the model performance.

\section{Conclusion}

We conduct a comprehensive evaluation across eight LLMs, focusing on their classification performance and generated explanations for detecting antisemitism.
We introduce an engineered CoT-like prompting, Guided-CoT, comparing its influence on model performance, generated explanations, and how effectively it handles in-context guidelines.
Additionally, our quantitative approach to assessing LLM-generated responses reveals paradoxical behavior in some models.
Through extensive experiments across various decoding strategies, prompting techniques, and the explanations they generate, we highlight key behaviors that may limit their utility and reliability, as well as their strengths and shortcomings, in a nuanced task like detecting antisemitism.
Future research can use these findings to audit models qualitatively, enhancing our understanding of the limitations associated with misclassification and potential biases.

\section{Limitations}

\textbf{Dataset.} The human-annotated dataset used in our study to design and evaluate the guided-CoT prompting technique has two primary limitations: 
First, it only encompasses the mainstream platform Twitter and tweets only in English, limiting the evaluation of our prompt design in a multilingual setting.
A more comprehensive evaluation may include platforms like Reddit~\citep{chen2024isamasred}, as well as alternatives, like Scored~\citep{patel2024idrama} and Lemmygrad~\citep{balci2025exploring}, where Israel- and Jewish-related topics frequently appear.
Additionally, multimodal contexts, specifically video-centric platforms like YouTube and Rumble~\citep{balci-etal-2025-podcast, balci2024idrama}, have also revealed discussions on this domain, providing researchers with an opportunity to investigate how models interpret moderation policies.
Second, the absolute performance of studied models may vary slightly, as annotators of the Twitter dataset were allowed to look at additional context during labeling, e.g., attached images, hyperlinks, replies, likes, and comments.
Upon qualitatively examining false negatives, we find cases where the models' explanations explicitly request more context, instances where the authors of this study also agree that additional context would be necessary, contradicting the ground truth, which is labeled antisemitic.

\textbf{Evaluation.} We evaluate our prompting strategies across three decoding strategies: 1) Greedy, 2) Sampling, and 3) Self-consistency.
However, we limit the hyperparameters for sampling and self-consistency to a single set, recognizing that alternative hyperparameter configurations could result in slight variations in the presented numbers.
Nonetheless, we run extensive self-consistency evaluations across all data points for six models, with each instance evaluated 30 times.
Since Self-consistency can reduce the measurement error-rate as we increase the inferences and are robust to prompt formats, our empirical findings about the comparison of models are robust as per Self-consistency.

We use UMAP for dimensionality reduction to reduce LLM-generated responses, specifically reducing the embeddings to 15 dimensions.
Selecting an appropriate dimensionality is challenging, as different models may benefit optimally from different dimensions.
We conduct a systematic grid search of UMAP hyperparameters to avoid arbitrary selection, selecting dimensionality based on maximizing the trustworthiness metric~\citep{pmlr-v5-maaten09a, trustworthiness}.
Moreover, we also back up our observations of LLM-generated responses with a significance test.

\textbf{Guided-CoT.} Our proposed Guided-CoT prompting shows benefits across evaluated models; however, the scope of our contribution is limited to eight models, two baseline prompting techniques (Zero-Shot and Zero-Shot-CoT), and one policy guideline (the IHRA definition).
Additionally, we only evaluate Guided-CoT on a single dataset collected from Twitter.
We acknowledge that the generalizability aspect of Guided-CoT is limited and under-explored for diverse datasets, multiple social media platforms, multilingual contexts, and different moderation policies.

Thoughts we include in the Guided-CoT are engineered for our task and might not generalize to other downstream tasks.
Our work primarily explores whether LLMs can effectively leverage in-context instructions for antisemitism detection.
Nonetheless, researchers can directly utilize the methodology and insights from this study for their work.

Guided-CoT using Llama 70B outperforms fine-tuned GPT-3.5, evaluated on about 6.9K data points.
Again, we rely on the numbers reported in ~\citet{becker2024antisemitism}'s study and do not have access to more details on when and how the model was trained.
Although the comparison lacks specific details, we aimed to measure a carefully engineered prompting technique versus a fine-tuned proprietary model, not to claim the superiority of our prompting, which could be generalizable for any task.

\textbf{Analyses on Subset of Dataset.} One of the challenges in evaluating and understanding the use of LLMs for detecting antisemitism is the safety-aligned release from developers.
During our initial experiments, we find that some models generate invalid responses that we cannot assess for our analysis.
Therefore, we slice the dataset into subsets for particular analyses to address this.
However, we treat each analysis independently to maximize the number of usable data points, and this may not be an ideal comparison.
Conversely, if we had used the same subset across the entire study, we would have been limited to approximately 62\% of the data.
We recognize this limitation and, as a result, do not compare results across different analysis setups.
For example, to answer the question, ``How do models perform under deterministic decoding?'' we hold subset-X constant across all eight models and report the findings accordingly.

\section{Ethical considerations}

This study examines antisemitism detection using the International Holocaust Remembrance Alliance's (IHRA) working definition~\citep{ihra-def}.
We emphasize that the authors did not develop this definition, but it is widely recognized and actively adopted by numerous governments and international bodies~\citep{ihra-members}.
For instance, the United States uses the IHRA's definition, highlighting its substantial real-world impact beyond academic contexts~\citep{ihra-usa}.
Our utilization of the IHRA definition does not reflect our judgment regarding its validity.
Instead, it highlights its practical significance and applicability in moderation policies implemented by platforms and governing bodies worldwide.

Our analyses are conducted with datasets annotated according to the IHRA definition by prior researchers~\citep{jikeli2023antisemitic, jikeli2024annotating}.
Any potential biases inherent to these datasets or annotations are comparable to those in other hate speech and toxicity datasets widely used in the literature.
We stress that our work does not involve subjective judgments by the authors regarding specific political or religious contexts beyond the explicit content of the IHRA definition and accompanying examples.

We affirm that our primary goal is to advance research on hate speech detection to facilitate more effective moderation systems.
We recognize the importance of cautious interpretation and application of our research findings in real-world scenarios, particularly given automated content moderation systems' complexities and potential societal impact.

\section*{Acknowledgments}

This work was supported by the NSF under Grants 2046590 and 2419831 as well as a gift from the Secunda Family Foundation.
We gratefully acknowledge use of the research computing resources of the Empire AI Consortium, Inc, with support from Empire State Development of the State of New York, the Simons Foundation, and the Secunda Family Foundation~\citep{Bloom2025EmpireAI}.

\bibliographystyle{ACM-Reference-Format}

\begin{thebibliography}{77}
\providecommand{\natexlab}[1]{#1}

\bibitem[{Aldreabi and Blackburn(2023)}]{aldreabi2023enhancing}
Esraa Aldreabi and Jeremy Blackburn. 2023.
\newblock Enhancing automated hate speech detection: Addressing islamophobia
  and freedom of speech in online discussions.
\newblock In \emph{Proceedings of the International Conference on Advances in
  Social Networks Analysis and Mining}, pages 644--651.

\bibitem[{AlKhamissi et~al.(2022)AlKhamissi, Ladhak, Iyer, Stoyanov, Kozareva,
  Li, Fung, Mathias, Celikyilmaz, and Diab}]{alkhamissi2022token}
Badr AlKhamissi, Faisal Ladhak, Srini Iyer, Ves Stoyanov, Zornitsa Kozareva,
  Xian Li, Pascale Fung, Lambert Mathias, Asli Celikyilmaz, and Mona Diab.
  2022.
\newblock Token: Task decomposition and knowledge infusion for few-shot hate
  speech detection.
\newblock \emph{arXiv preprint arXiv:2205.12495}.

\bibitem[{Antypas and Camacho-Collados(2023)}]{antypas2023robust}
Dimosthenis Antypas and Jose Camacho-Collados. 2023.
\newblock Robust hate speech detection in social media: A cross-dataset
  empirical evaluation.
\newblock \emph{arXiv preprint arXiv:2307.01680}.

\bibitem[{Atreja et~al.(2024)Atreja, Ashkinaze, Li, Mendelsohn, and
  Hemphill}]{atreja2024prompt}
Shubham Atreja, Joshua Ashkinaze, Lingyao Li, Julia Mendelsohn, and Libby
  Hemphill. 2024.
\newblock Prompt design matters for computational social science tasks but in
  unpredictable ways.
\newblock \emph{arXiv preprint arXiv:2406.11980}.

\bibitem[{Balachandran et~al.(2024)Balachandran, Chen, Joshi, Nushi, Palangi,
  Salinas, Vineet, Woffinden-Luey, and Yousefi}]{balachandran2024eureka}
Vidhisha Balachandran, Jingya Chen, Neel Joshi, Besmira Nushi, Hamid Palangi,
  Eduardo Salinas, Vibhav Vineet, James Woffinden-Luey, and Safoora Yousefi.
  2024.
\newblock Eureka: Evaluating and understanding large foundation models.
\newblock \emph{arXiv preprint arXiv:2409.10566}.

\bibitem[{Balci et~al.(2024)Balci, Patel, Balci, and
  Blackburn}]{balci2024idrama}
Utkucan Balci, Jay Patel, Berkan Balci, and Jeremy Blackburn. 2024.
\newblock idrama-rumble-2024: A dataset of podcasts from rumble spanning 2020
  to 2022.
\newblock In \emph{Workshop Proceedings of the 18th International AAAI
  Conference on Web and Social Media}.

\bibitem[{Balci et~al.(2025{\natexlab{a}})Balci, Patel, Balci, and
  Blackburn}]{balci-etal-2025-podcast}
Utkucan Balci, Jay Patel, Berkan Balci, and Jeremy Blackburn.
  2025{\natexlab{a}}.
\newblock \href {https://doi.org/10.18653/v1/2025.nlp4dh-1.6} {Podcast
  outcasts: Understanding rumble{'}s podcast dynamics}.
\newblock In \emph{Proceedings of the 5th International Conference on Natural
  Language Processing for Digital Humanities}, pages 48--62, Albuquerque, USA.
  Association for Computational Linguistics.

\bibitem[{Balci et~al.(2025{\natexlab{b}})Balci, Sirivianos, and
  Blackburn}]{balci2025exploring}
Utkucan Balci, Michael Sirivianos, and Jeremy Blackburn. 2025{\natexlab{b}}.
\newblock Exploring left-wing extremism on the decentralized web: An analysis
  of lemmygrad. ml.
\newblock \emph{arXiv preprint arXiv:2507.23699}.

\bibitem[{Becker et~al.(2024)Becker, Ascone, Placzynta, and
  Vincent}]{becker2024antisemitism}
Matthias~Jakob Becker, Laura Ascone, Karolina Placzynta, and Chlo{\'e} Vincent.
  2024.
\newblock Antisemitism in online communication: Transdisciplinary approaches to
  hate speech in the twenty-first century.

\bibitem[{BehnamGhader et~al.(2024)BehnamGhader, Adlakha, Mosbach, Bahdanau,
  Chapados, and Reddy}]{behnamghader2024llm2vec}
Parishad BehnamGhader, Vaibhav Adlakha, Marius Mosbach, Dzmitry Bahdanau,
  Nicolas Chapados, and Siva Reddy. 2024.
\newblock Llm2vec: Large language models are secretly powerful text encoders.
\newblock \emph{arXiv preprint arXiv:2404.05961}.

\bibitem[{Bloom et~al.(2025)Bloom, Brumberg, Fisk, Harrison, Hull,
  Ramasubramanian, Vliet, and Wing}]{Bloom2025EmpireAI}
Stacie Bloom, Joshua~C. Brumberg, Ian Fisk, Robert~J. Harrison, Robert Hull,
  Melur Ramasubramanian, Krystyn~Van Vliet, and Jeannette Wing. 2025.
\newblock \href {https://doi.org/10.1145/3708035.3736070} {Empire {AI}: A new
  model for provisioning {AI} and {HPC} for academic research in the public
  good}.
\newblock In \emph{Practice and Experience in Advanced Research Computing
  ({PEARC} '25)}, page~4, Columbus, OH, USA. ACM.

\bibitem[{Casula and Tonelli(2024)}]{casula2024target}
Camilla Casula and Sara Tonelli. 2024.
\newblock A target-aware analysis of data augmentation for hate speech
  detection.
\newblock \emph{arXiv preprint arXiv:2410.08053}.

\bibitem[{Chen et~al.(2024{\natexlab{a}})Chen, Shen, Bavalatti, Lin, Wang, Hu,
  Subramanyam, Vepuri, Jiang, Qi et~al.}]{chen2024class}
Jianfa Chen, Emily Shen, Trupti Bavalatti, Xiaowen Lin, Yongkai Wang, Shuming
  Hu, Harihar Subramanyam, Ksheeraj~Sai Vepuri, Ming Jiang, Ji~Qi, et~al.
  2024{\natexlab{a}}.
\newblock Class-rag: Real-time content moderation with retrieval augmented
  generation.
\newblock \emph{arXiv preprint arXiv:2410.14881}.

\bibitem[{Chen et~al.(2024{\natexlab{b}})Chen, He, Burghardt, Zhang, and
  Lerman}]{chen2024isamasred}
Kai Chen, Zihao He, Keith Burghardt, Jingxin Zhang, and Kristina Lerman.
  2024{\natexlab{b}}.
\newblock Isamasred: A public dataset tracking reddit discussions on
  israel-hamas conflict.
\newblock In \emph{Proceedings of the International AAAI Conference on Web and
  Social Media}, volume~18, pages 1900--1912.

\bibitem[{Common(2020)}]{common2020fear}
MacKenzie~F Common. 2020.
\newblock Fear the reaper: How content moderation rules are enforced on social
  media.
\newblock \emph{International Review of Law, Computers \& Technology},
  34(2):126--152.

\bibitem[{Di~Bonaventura et~al.(2024)Di~Bonaventura, Siciliani, Basile,
  Mero{\~n}o-Pe{\~n}uela, and McGillivray}]{di2024explanation}
Chiara Di~Bonaventura, Lucia Siciliani, Pierpaolo Basile, Albert
  Mero{\~n}o-Pe{\~n}uela, and Barbara McGillivray. 2024.
\newblock Is explanation all you need? an expert survey on llm-generated
  explanations for abusive language detection.
\newblock In \emph{Proceedings of the 10th Italian Conference on Computational
  Linguistics (CLiC-it 2024)}, pages 280--288.

\bibitem[{Dutta et~al.(2024)Dutta, Khorramrouz, Dutta, and
  KhudaBukhsh}]{dutta2024down}
Arka Dutta, Adel Khorramrouz, Sujan Dutta, and Ashiqur~R KhudaBukhsh. 2024.
\newblock Down the toxicity rabbit hole: A framework to bias audit large
  language models with key emphasis on racism, antisemitism, and misogyny.
\newblock In \emph{Proceedings of the Thirty-Third International Joint
  Conference on Artificial Intelligence, IJCAI}, pages 3--9.

\bibitem[{Fasching and Lelkes(2025)}]{fasching2025model}
Neil Fasching and Yphtach Lelkes. 2025.
\newblock Model-dependent moderation: Inconsistencies in hate speech detection
  across llm-based systems.
\newblock In \emph{Findings of the Association for Computational Linguistics:
  ACL 2025}, pages 22271--22285.

\bibitem[{Goldberg et~al.(2024)Goldberg, Acosta-Navas, Bakker, Beacock,
  Botvinick, Buch, DiResta, Donthi, Fast, Iyer et~al.}]{goldberg2024ai}
Beth Goldberg, Diana Acosta-Navas, Michiel Bakker, Ian Beacock, Matt Botvinick,
  Prateek Buch, Ren{\'e}e DiResta, Nandika Donthi, Nathanael Fast, Ravi Iyer,
  et~al. 2024.
\newblock Ai and the future of digital public squares.
\newblock \emph{arXiv preprint arXiv:2412.09988}.

\bibitem[{Goyal et~al.(2025)Goyal, Zhan, Chen, Saha, and
  Chandrasekharan}]{goyal2025momoe}
Agam Goyal, Xianyang Zhan, Yilun Chen, Koustuv Saha, and Eshwar
  Chandrasekharan. 2025.
\newblock Momoe: Mixture of moderation experts framework for ai-assisted online
  governance.
\newblock \emph{arXiv preprint arXiv:2505.14483}.

\bibitem[{Guo et~al.(2023)Guo, Hu, Mu, Shi, Zhao, Vishwamitra, and
  Hu}]{guo2023investigation}
Keyan Guo, Alexander Hu, Jaden Mu, Ziheng Shi, Ziming Zhao, Nishant
  Vishwamitra, and Hongxin Hu. 2023.
\newblock An investigation of large language models for real-world hate speech
  detection.
\newblock In \emph{2023 International Conference on Machine Learning and
  Applications (ICMLA)}, pages 1568--1573. IEEE.

\bibitem[{Halterman and Keith(2024)}]{halterman2024codebook}
Andrew Halterman and Katherine~A Keith. 2024.
\newblock Codebook llms: Adapting political science codebooks for llm use and
  adapting llms to follow codebooks.
\newblock \emph{arXiv preprint arXiv:2407.10747}.

\bibitem[{Hanshaw(2025)}]{hanshawCriticsRaiseFree}
Annelise Hanshaw. 2025.
\newblock Critics raise free speech concerns as {{Missouri House}} advances
  bill targeting antisemitism.
\newblock
  https://www.news-leader.com/story/news/politics/2025/04/08/missouri-house-bill-targeting-antisemitism-prompts-free-speech-worries/82990192007/.

\bibitem[{Hartvigsen et~al.(2022)Hartvigsen, Gabriel, Palangi, Sap, Ray, and
  Kamar}]{hartvigsen2022toxigen}
Thomas Hartvigsen, Saadia Gabriel, Hamid Palangi, Maarten Sap, Dipankar Ray,
  and Ece Kamar. 2022.
\newblock Toxigen: A large-scale machine-generated dataset for adversarial and
  implicit hate speech detection.
\newblock \emph{arXiv preprint arXiv:2203.09509}.

\bibitem[{Huang(2025)}]{huang2025content}
Tao Huang. 2025.
\newblock Content moderation by llm: From accuracy to legitimacy.
\newblock \emph{Artificial Intelligence Review}, 58(10):1--32.

\bibitem[{IHRA(2024)}]{ihra-def}
IHRA. 2024.
\newblock Working definition of antisemitism.
\newblock
  \url{https://holocaustremembrance.com/resources/working-definition-antisemitism}.
\newblock Accessed: 2024-12-15.

\bibitem[{IHRA-MEMBERS(2024)}]{ihra-members}
IHRA-MEMBERS. 2024.
\newblock The ihra member countries.
\newblock \url{https://holocaustremembrance.com/who-we-are/member-countries}.
\newblock Accessed: 2024-12-15.

\bibitem[{IHRA-USA(2024)}]{ihra-usa}
IHRA-USA. 2024.
\newblock Defining antisemitism.
\newblock \url{https://www.state.gov/defining-antisemitism/}.
\newblock Accessed: 2024-12-15.

\bibitem[{Jahan et~al.(2024)Jahan, Oussalah, Beddia, Arhab
  et~al.}]{jahan2024comprehensive}
Md~Saroar Jahan, Mourad Oussalah, Djamila~Romaissa Beddia, Nabil Arhab, et~al.
  2024.
\newblock A comprehensive study on nlp data augmentation for hate speech
  detection: Legacy methods, bert, and llms.
\newblock \emph{arXiv preprint arXiv:2404.00303}.

\bibitem[{Jikeli et~al.(2023{\natexlab{a}})Jikeli, Karali, Miehling, and
  Soemer}]{jikeli2023antisemitic}
Gunther Jikeli, Sameer Karali, Daniel Miehling, and Katharina Soemer.
  2023{\natexlab{a}}.
\newblock Antisemitic messages? a guide to high-quality annotation and a
  labeled dataset of tweets.
\newblock \emph{arXiv preprint arXiv:2304.14599}.

\bibitem[{Jikeli et~al.(2023{\natexlab{b}})Jikeli, Karali, Miehling, and
  Soemer}]{ihra-dataset-v1}
Gunther Jikeli, Sameer Karali, Daniel Miehling, and Katharina Soemer.
  2023{\natexlab{b}}.
\newblock \href {https://zenodo.org/records/7872835} {[version 1] antisemitism
  on twitter: A dataset for machine learning and text analytics}.
\newblock Accessed on Jan. 17, 2024.

\bibitem[{Jikeli et~al.(2024{\natexlab{a}})Jikeli, Karali, Miehling, and
  Soemer}]{ihra-dataset-v3}
Gunther Jikeli, Sameer Karali, Daniel Miehling, and Katharina Soemer.
  2024{\natexlab{a}}.
\newblock \href {https://zenodo.org/records/14448399} {[version 3] antisemitism
  on twitter: A dataset for machine learning and text analytics}.
\newblock Accessed on Jan. 17, 2024.

\bibitem[{Jikeli et~al.(2024{\natexlab{b}})Jikeli, Soemer, and
  Karali}]{jikeli2024annotating}
Gunther Jikeli, Katharina Soemer, and Sameer Karali. 2024{\natexlab{b}}.
\newblock Annotating live messages on social media. testing the efficiency of
  the annothate--live data annotation portal.
\newblock \emph{Journal of Computational Social Science}, 7(1):571--585.

\bibitem[{Just et~al.(2025)Just, Dabas, Huang, Jin, and
  Jia}]{just-etal-2025-dipt}
Hoang~Anh Just, Mahavir Dabas, Lifu Huang, Ming Jin, and Ruoxi Jia. 2025.
\newblock \href {https://aclanthology.org/2025.findings-naacl.356/} {{D}i{PT}:
  Enhancing {LLM} reasoning through diversified perspective-taking}.
\newblock In \emph{Findings of the Association for Computational Linguistics:
  NAACL 2025}, pages 6344--6374, Albuquerque, New Mexico. Association for
  Computational Linguistics.

\bibitem[{Kikkisetti et~al.(2024)Kikkisetti, Mustafa, Melillo, Corizzo,
  Boukouvalas, Gill, and Japkowicz}]{kikkisetti2024using}
Dhanush Kikkisetti, Raza~Ul Mustafa, Wendy Melillo, Roberto Corizzo, Zois
  Boukouvalas, Jeff Gill, and Nathalie Japkowicz. 2024.
\newblock Using llms to discover emerging coded antisemitic hate-speech in
  extremist social media.
\newblock \emph{arXiv preprint arXiv:2401.10841}.

\bibitem[{Kojima et~al.(2022)Kojima, Gu, Reid, Matsuo, and
  Iwasawa}]{kojima2022large}
Takeshi Kojima, Shixiang~Shane Gu, Machel Reid, Yutaka Matsuo, and Yusuke
  Iwasawa. 2022.
\newblock Large language models are zero-shot reasoners.
\newblock \emph{Advances in neural information processing systems},
  35:22199--22213.

\bibitem[{Kolla et~al.(2024)Kolla, Salunkhe, Chandrasekharan, and
  Saha}]{kolla2024llm}
Mahi Kolla, Siddharth Salunkhe, Eshwar Chandrasekharan, and Koustuv Saha. 2024.
\newblock Llm-mod: Can large language models assist content moderation?
\newblock In \emph{Extended Abstracts of the CHI Conference on Human Factors in
  Computing Systems}, pages 1--8.

\bibitem[{Kumar et~al.(2024{\natexlab{a}})Kumar, AbuHashem, and
  Durumeric}]{kumar2024watch}
Deepak Kumar, Yousef~Anees AbuHashem, and Zakir Durumeric. 2024{\natexlab{a}}.
\newblock Watch your language: Investigating content moderation with large
  language models.
\newblock In \emph{Proceedings of the International AAAI Conference on Web and
  Social Media}, volume~18, pages 865--878.

\bibitem[{Kumar et~al.(2024{\natexlab{b}})Kumar, Kholkar, Mendke, Sadana,
  Agrawal, and Dandapat}]{kumar2024socio}
Shanu Kumar, Gauri Kholkar, Saish Mendke, Anubhav Sadana, Parag Agrawal, and
  Sandipan Dandapat. 2024{\natexlab{b}}.
\newblock Socio-culturally aware evaluation framework for llm-based content
  moderation.
\newblock \emph{arXiv preprint arXiv:2412.13578}.

\bibitem[{Kumarage et~al.(2024)Kumarage, Bhattacharjee, and
  Garland}]{kumarage2024harnessing}
Tharindu Kumarage, Amrita Bhattacharjee, and Joshua Garland. 2024.
\newblock Harnessing artificial intelligence to combat online hate: Exploring
  the challenges and opportunities of large language models in hate speech
  detection.
\newblock \emph{arXiv preprint arXiv:2403.08035}.

\bibitem[{Li et~al.(2023)Li, Wang, Zhu, Zhang, Hou, Lian, and
  Xie}]{li2023emotionprompt}
Cheng Li, Jindong Wang, Kaijie Zhu, Yixuan Zhang, Wenxin Hou, Jianxun Lian, and
  Xing Xie. 2023.
\newblock Emotionprompt: Leveraging psychology for large language models
  enhancement via emotional stimulus.
\newblock \emph{arXiv e-prints}, pages arXiv--2307.

\bibitem[{Liao et~al.(2025)Liao, Hu, Zhu, He, and Jin}]{liao-etal-2025-forest}
Haoran Liao, Shaohua Hu, Zhihao Zhu, Hao He, and Yaohui Jin. 2025.
\newblock \href {https://aclanthology.org/2025.naacl-long.66/} {Forest for the
  trees: Overarching prompting evokes high-level reasoning in large language
  models}.
\newblock In \emph{Proceedings of the 2025 Conference of the Nations of the
  Americas Chapter of the Association for Computational Linguistics: Human
  Language Technologies (Volume 1: Long Papers)}, pages 1433--1453,
  Albuquerque, New Mexico. Association for Computational Linguistics.

\bibitem[{Lin et~al.(2023)Lin, Wang, Tong, Wang, Guo, Wang, and
  Shang}]{lin2023toxicchat}
Zi~Lin, Zihan Wang, Yongqi Tong, Yangkun Wang, Yuxin Guo, Yujia Wang, and
  Jingbo Shang. 2023.
\newblock Toxicchat: Unveiling hidden challenges of toxicity detection in
  real-world user-ai conversation.
\newblock \emph{arXiv preprint arXiv:2310.17389}.

\bibitem[{Ling et~al.(2022)Ling, Blackburn, De~Cristofaro, and
  Stringhini}]{ling2022slapping}
Chen Ling, Jeremy Blackburn, Emiliano De~Cristofaro, and Gianluca Stringhini.
  2022.
\newblock Slapping cats, bopping heads, and oreo shakes: Understanding
  indicators of virality in tiktok short videos.
\newblock In \emph{Proceedings of the 14th ACM Web Science Conference 2022},
  pages 164--173.

\bibitem[{Mathew et~al.(2021)Mathew, Saha, Yimam, Biemann, Goyal, and
  Mukherjee}]{mathew2021hatexplain}
Binny Mathew, Punyajoy Saha, Seid~Muhie Yimam, Chris Biemann, Pawan Goyal, and
  Animesh Mukherjee. 2021.
\newblock Hatexplain: A benchmark dataset for explainable hate speech
  detection.
\newblock In \emph{Proceedings of the AAAI conference on artificial
  intelligence}, volume~35, pages 14867--14875.

\bibitem[{McInnes et~al.(2018)McInnes, Healy, and Melville}]{mcinnes2018umap}
Leland McInnes, John Healy, and James Melville. 2018.
\newblock Umap: Uniform manifold approximation and projection for dimension
  reduction.
\newblock \emph{arXiv preprint arXiv:1802.03426}.

\bibitem[{Mekala et~al.(2023)Mekala, Razeghi, and Singh}]{mekala2023echoprompt}
Rajasekhar~Reddy Mekala, Yasaman Razeghi, and Sameer Singh. 2023.
\newblock Echoprompt: Instructing the model to rephrase queries for improved
  in-context learning.
\newblock \emph{arXiv preprint arXiv:2309.10687}.

\bibitem[{Moses(2025)}]{mosesYaleAddsContested2025}
Nora Ada~Perlman Moses, Josie~Reich. 2025.
\newblock Yale adds contested antisemitism definition to discrimination policy.
\newblock
  https://yaledailynews.com/blog/2025/04/08/yale-adds-contested-antisemitism-definition-to-discrimination-policy/.

\bibitem[{Nghiem and Daum{\'e}~III(2024)}]{nghiem2024hatecot}
Huy Nghiem and Hal Daum{\'e}~III. 2024.
\newblock Hatecot: An explanation-enhanced dataset for generalizable offensive
  speech detection via large language models.
\newblock \emph{arXiv preprint arXiv:2403.11456}.

\bibitem[{Nirmal et~al.(2024)Nirmal, Bhattacharjee, Sheth, and
  Liu}]{nirmal2024towards}
Ayushi Nirmal, Amrita Bhattacharjee, Paras Sheth, and Huan Liu. 2024.
\newblock Towards interpretable hate speech detection using large language
  model-extracted rationales.
\newblock \emph{arXiv preprint arXiv:2403.12403}.

\bibitem[{Palla et~al.(2025)Palla, Garc{\'\i}a, Hauff, Fabbri, Lindstr{\"o}m,
  Taber, Damianou, and Lalmas}]{palla2025policy}
Konstantina Palla, Jos{\'e} Luis~Redondo Garc{\'\i}a, Claudia Hauff, Francesco
  Fabbri, Henrik Lindstr{\"o}m, Daniel~R Taber, Andreas Damianou, and Mounia
  Lalmas. 2025.
\newblock Policy-as-prompt: Rethinking content moderation in the age of large
  language models.
\newblock \emph{arXiv preprint arXiv:2502.18695}.

\bibitem[{Patel et~al.(2024)Patel, Paudel, De~Cristofaro, Stringhini, and
  Blackburn}]{patel2024idrama}
Jay Patel, Pujan Paudel, Emiliano De~Cristofaro, Gianluca Stringhini, and
  Jeremy Blackburn. 2024.
\newblock \href {https://doi.org/10.1609/icwsm.v18i1.31444}
  {idrama-scored-2024: A dataset of the scored social media platform from 2020
  to 2023}.
\newblock In \emph{Proceedings of the International AAAI Conference on Web and
  Social Media}, volume~18, pages 2014--2024.

\bibitem[{Pedregosa et~al.(2011)Pedregosa, Varoquaux, Gramfort, Michel,
  Thirion, Grisel, Blondel, Prettenhofer, Weiss, Dubourg, Vanderplas, Passos,
  Cournapeau, Brucher, Perrot, and Duchesnay}]{trustworthiness}
F.~Pedregosa, G.~Varoquaux, A.~Gramfort, V.~Michel, B.~Thirion, O.~Grisel,
  M.~Blondel, P.~Prettenhofer, R.~Weiss, V.~Dubourg, J.~Vanderplas, A.~Passos,
  D.~Cournapeau, M.~Brucher, M.~Perrot, and E.~Duchesnay. 2011.
\newblock \href
  {https://scikit-learn.org/stable/modules/generated/sklearn.manifold.trustworthiness.html}
  {Trustworthiness: Scikit-learn: Machine learning in {P}ython}.
\newblock Accessed on Jan. 17, 2024.

\bibitem[{Piot and Parapar(2024)}]{piot2024decoding}
Paloma Piot and Javier Parapar. 2024.
\newblock Decoding hate: Exploring language models' reactions to hate speech.
\newblock \emph{arXiv preprint arXiv:2410.00775}.

\bibitem[{Renze and Guven(2024)}]{renze2024effect}
Matthew Renze and Erhan Guven. 2024.
\newblock The effect of sampling temperature on problem solving in large
  language models.
\newblock \emph{arXiv preprint arXiv:2402.05201}.

\bibitem[{Roy et~al.(2023)Roy, Harshavardhan, Mukherjee, and
  Saha}]{roy2023probing}
Sarthak Roy, Ashish Harshavardhan, Animesh Mukherjee, and Punyajoy Saha. 2023.
\newblock Probing llms for hate speech detection: strengths and
  vulnerabilities.
\newblock \emph{arXiv preprint arXiv:2310.12860}.

\bibitem[{Sap et~al.(2019)Sap, Card, Gabriel, Choi, and Smith}]{sap2019risk}
Maarten Sap, Dallas Card, Saadia Gabriel, Yejin Choi, and Noah~A Smith. 2019.
\newblock The risk of racial bias in hate speech detection.
\newblock In \emph{Proceedings of the 57th annual meeting of the association
  for computational linguistics}, pages 1668--1678.

\bibitem[{Schulhoff et~al.(2024)Schulhoff, Ilie, Balepur, Kahadze, Liu, Si, Li,
  Gupta, Han, Schulhoff, Dulepet, Vidyadhara, Ki, Agrawal, Pham, Kroiz, Li,
  Tao, Srivastava, Costa, Gupta, Rogers, Goncearenco, Sarli, Galynker, Peskoff,
  Carpuat, White, Anadkat, Hoyle, and Resnik}]{schulhoff2024prompt}
Sander Schulhoff, Michael Ilie, Nishant Balepur, Konstantine Kahadze, Amanda
  Liu, Chenglei Si, Yinheng Li, Aayush Gupta, HyoJung Han, Sevien Schulhoff,
  Pranav~Sandeep Dulepet, Saurav Vidyadhara, Dayeon Ki, Sweta Agrawal, Chau
  Pham, Gerson Kroiz, Feileen Li, Hudson Tao, Ashay Srivastava, Hevander~Da
  Costa, Saloni Gupta, Megan~L. Rogers, Inna Goncearenco, Giuseppe Sarli, Igor
  Galynker, Denis Peskoff, Marine Carpuat, Jules White, Shyamal Anadkat,
  Alexander Hoyle, and Philip Resnik. 2024.
\newblock \href {https://arxiv.org/abs/2406.06608} {The prompt report: A
  systematic survey of prompting techniques}.
\newblock \emph{Preprint}, arXiv:2406.06608.

\bibitem[{Sclar et~al.(2023)Sclar, Choi, Tsvetkov, and
  Suhr}]{sclar2023quantifying}
Melanie Sclar, Yejin Choi, Yulia Tsvetkov, and Alane Suhr. 2023.
\newblock Quantifying language models' sensitivity to spurious features in
  prompt design or: How i learned to start worrying about prompt formatting.
\newblock \emph{arXiv preprint arXiv:2310.11324}.

\bibitem[{Shaikh et~al.(2022)Shaikh, Zhang, Held, Bernstein, and
  Yang}]{shaikh2022second}
Omar Shaikh, Hongxin Zhang, William Held, Michael Bernstein, and Diyi Yang.
  2022.
\newblock On second thought, let's not think step by step! bias and toxicity in
  zero-shot reasoning.
\newblock \emph{arXiv preprint arXiv:2212.08061}.

\bibitem[{Smith(2025)}]{smithHarvardAgreesAdopt2025}
Tovia Smith. 2025.
\newblock Harvard agrees to adopt a broad definition of antisemitism to settle
  two lawsuits.
\newblock \emph{NPR}.

\bibitem[{Song et~al.(2024)Song, Wang, Li, and Lin}]{song2024good}
Yifan Song, Guoyin Wang, Sujian Li, and Bill~Yuchen Lin. 2024.
\newblock The good, the bad, and the greedy: Evaluation of llms should not
  ignore non-determinism.
\newblock \emph{arXiv preprint arXiv:2407.10457}.

\bibitem[{Tahmasbi et~al.(2021)Tahmasbi, Schild, Ling, Blackburn, Stringhini,
  Zhang, and Zannettou}]{tahmasbiGoEatBat2021}
Fatemeh Tahmasbi, Leonard Schild, Chen Ling, Jeremy Blackburn, Gianluca
  Stringhini, Yang Zhang, and Savvas Zannettou. 2021.
\newblock \href {https://doi.org/10.1145/3442381.3450024} {``{{Go}} eat a bat,
  chang!'': {{On}} the emergence of sinophobic behavior on web communities in
  the face of {{COVID-19}}}.
\newblock In \emph{Proceedings of the Web Conference 2021}, {{WWW}} '21, pages
  1122--1133, New York, NY, USA. Association for Computing Machinery.

\bibitem[{Tam et~al.(2024)Tam, Wu, Tsai, Lin, Lee, and Chen}]{tam2024let}
Zhi~Rui Tam, Cheng-Kuang Wu, Yi-Lin Tsai, Chieh-Yen Lin, Hung-yi Lee, and
  Yun-Nung Chen. 2024.
\newblock Let me speak freely? a study on the impact of format restrictions on
  performance of large language models.
\newblock \emph{arXiv preprint arXiv:2408.02442}.

\bibitem[{Thapa et~al.(2025)Thapa, Shiwakoti, Shah, Adhikari, Veeramani, Nasim,
  and Naseem}]{thapa2025large}
Surendrabikram Thapa, Shuvam Shiwakoti, Siddhant~Bikram Shah, Surabhi Adhikari,
  Hariram Veeramani, Mehwish Nasim, and Usman Naseem. 2025.
\newblock Large language models (llm) in computational social science:
  prospects, current state, and challenges.
\newblock \emph{Social Network Analysis and Mining}, 15(1):1--30.

\bibitem[{van~der Maaten(2009)}]{pmlr-v5-maaten09a}
Laurens van~der Maaten. 2009.
\newblock \href {https://proceedings.mlr.press/v5/maaten09a.html} {Learning a
  parametric embedding by preserving local structure}.
\newblock In \emph{Proceedings of the Twelfth International Conference on
  Artificial Intelligence and Statistics}, volume~5 of \emph{Proceedings of
  Machine Learning Research}, pages 384--391, Hilton Clearwater Beach Resort,
  Clearwater Beach, Florida USA. PMLR.

\bibitem[{Vijjini et~al.(2025)Vijjini, Basu Roy~Chowdhury, and
  Chaturvedi}]{vijjini-etal-2025-exploring}
Anvesh~Rao Vijjini, Somnath Basu Roy~Chowdhury, and Snigdha Chaturvedi. 2025.
\newblock \href {https://aclanthology.org/2025.naacl-long.565/} {Exploring
  safety-utility trade-offs in personalized language models}.
\newblock In \emph{Proceedings of the 2025 Conference of the Nations of the
  Americas Chapter of the Association for Computational Linguistics: Human
  Language Technologies (Volume 1: Long Papers)}, pages 11316--11340,
  Albuquerque, New Mexico. Association for Computational Linguistics.

\bibitem[{Vishwamitra et~al.(2024)Vishwamitra, Guo, Romit, Ondracek, Cheng,
  Zhao, and Hu}]{vishwamitra2024moderating}
Nishant Vishwamitra, Keyan Guo, Farhan~Tajwar Romit, Isabelle Ondracek, Long
  Cheng, Ziming Zhao, and Hongxin Hu. 2024.
\newblock Moderating new waves of online hate with chain-of-thought reasoning
  in large language models.
\newblock In \emph{2024 IEEE Symposium on Security and Privacy (SP)}, pages
  788--806. IEEE.

\bibitem[{Wan et~al.(2024)Wan, Wu, Chen, and Li}]{wan2024reasoning}
Guangya Wan, Yuqi Wu, Jie Chen, and Sheng Li. 2024.
\newblock Reasoning aware self-consistency: Leveraging reasoning paths for
  efficient llm sampling.
\newblock \emph{arXiv preprint arXiv:2408.17017}.

\bibitem[{Wang et~al.(2022)Wang, Wei, Schuurmans, Le, Chi, Narang, Chowdhery,
  and Zhou}]{wang2022self}
Xuezhi Wang, Jason Wei, Dale Schuurmans, Quoc Le, Ed~Chi, Sharan Narang,
  Aakanksha Chowdhery, and Denny Zhou. 2022.
\newblock Self-consistency improves chain of thought reasoning in language
  models.
\newblock \emph{arXiv preprint arXiv:2203.11171}.

\bibitem[{Wei et~al.(2022)Wei, Wang, Schuurmans, Bosma, Xia, Chi, Le, Zhou
  et~al.}]{wei2022chain}
Jason Wei, Xuezhi Wang, Dale Schuurmans, Maarten Bosma, Fei Xia, Ed~Chi, Quoc~V
  Le, Denny Zhou, et~al. 2022.
\newblock Chain-of-thought prompting elicits reasoning in large language
  models.
\newblock \emph{Advances in neural information processing systems},
  35:24824--24837.

\bibitem[{Yang et~al.(2023)Yang, Kim, Kim, Ho, Thorne, and Yun}]{yang2023hare}
Yongjin Yang, Joonkee Kim, Yujin Kim, Namgyu Ho, James Thorne, and Se-Young
  Yun. 2023.
\newblock Hare: Explainable hate speech detection with step-by-step reasoning.
\newblock \emph{arXiv preprint arXiv:2311.00321}.

\bibitem[{Yin et~al.(2025)Yin, Laban, Peng, Zhou, Mao, Vats, Ross, Agarwal,
  Xiong, and Wu}]{yin2025bingoguard}
Fan Yin, Philippe Laban, Xiangyu Peng, Yilun Zhou, Yixin Mao, Vaibhav Vats,
  Linnea Ross, Divyansh Agarwal, Caiming Xiong, and Chien-Sheng Wu. 2025.
\newblock Bingoguard: Llm content moderation tools with risk levels.
\newblock \emph{arXiv preprint arXiv:2503.06550}.

\bibitem[{Zannettou et~al.(2020)Zannettou, Finkelstein, Bradlyn, and
  Blackburn}]{zannettou2020quantitative}
Savvas Zannettou, Joel Finkelstein, Barry Bradlyn, and Jeremy Blackburn. 2020.
\newblock A quantitative approach to understanding online antisemitism.
\newblock In \emph{Proceedings of the International AAAI conference on Web and
  Social Media}, volume~14, pages 786--797.

\bibitem[{Zhang et~al.(2024)Zhang, He, Ji, and Lu}]{zhang2024don}
Min Zhang, Jianfeng He, Taoran Ji, and Chang-Tien Lu. 2024.
\newblock Don't go to extremes: Revealing the excessive sensitivity and
  calibration limitations of llms in implicit hate speech detection.
\newblock \emph{arXiv preprint arXiv:2402.11406}.

\bibitem[{Zhang et~al.(2025)Zhang, Xu, Wu, and Reddy}]{zhang2025falsereject}
Zhehao Zhang, Weijie Xu, Fanyou Wu, and Chandan~K Reddy. 2025.
\newblock Falsereject: A resource for improving contextual safety and
  mitigating over-refusals in llms via structured reasoning.
\newblock \emph{arXiv preprint arXiv:2505.08054}.

\bibitem[{Zheng et~al.(2024)Zheng, Liu, Haque, Qian, Yang, and
  Yang}]{zheng2024hatemoderate}
Jiangrui Zheng, Xueqing Liu, Mirazul Haque, Xing Qian, Guanqun Yang, and Wei
  Yang. 2024.
\newblock Hatemoderate: Testing hate speech detectors against content
  moderation policies.
\newblock In \emph{Findings of the Association for Computational Linguistics:
  NAACL 2024}, pages 2691--2710.

\end{thebibliography}

%
%

\newpage

\appendix

\addtocontents{toc}{\protect\setcounter{tocdepth}{2}}

\appendixtableofcontents

\begin{table*}[t]
\small
\centering
\begin{tabular}{l c c c c c c l}
\toprule
\textbf{Model} & \textbf{Size} & \textbf{Release Time} & \textbf{Multilingual} & \textbf{Reasoning} & \textbf{Quantized} & \textbf{Developer} \\
\midrule
Mistral 2410 & 8B  & Oct, 2024  & Yes & No  & No  & Mistral AI \\
Mistral 2501 & 24B & Jan, 2025  & Yes & No  & No  & Mistral AI \\
Gemma 3 & 12B  & Mar, 2025  & Yes  & No  & No  & Google \\
Gemma 3 & 27B & Mar, 2025  & Yes  & No  & No  & Google \\
Llama 3.1 & 8B  & Jul, 2024  & Yes & No  & No  & Meta \\
Llama 3.1 & 70B & Jul, 2024  & Yes & No  & Yes & Meta \\
DS-R1-Llama & 70B & Jan, 2025  & Yes & Yes & Yes & DeepSeek \\
QwQ & 32B & Mar, 2025 & Yes & Yes & No  & Qwen \\
\bottomrule
\end{tabular}
\caption{Models being evaluated in this study with their attributes.}
\label{apx:model-attributes}
\end{table*}

\section{Take-aways}
\label{apx:take-aways}

In this section, through our experiments, we consolidate all the differences observed as takeaways.

\subsection{Utility}

\begin{itemize}

    \item Different models exhibit distinct patterns of invalid responses.
    
    \item Gemma-family models generate the highest number of ambiguous classifications rather than providing explicit Yes/No labels.
    
    \item Llama 8B exhibits the highest refusal rate among all evaluated models. This behavior is potentially due to differences in their post-alignment training process.
    
    \item QwQ 32B often gets stuck in repetitive generation under greedy decoding.

    \item Guided CoT reduces Llama 8B's refusal rate to nearly Zero percent and lowers ambiguous responses from Gemma-family models.

    \item For all the models, important thoughts to improving antisemitism detection performance include: 
          1) explicitly mention checking for sarcasm and criticism,  
          2) explicitly instruct to check whether a post promotes antisemitism or spreads awareness, and 
          3) including a guiding step that encourages the model to reflect on analyzing whether the author's overall stance is favorable toward the Jewish community.
    
\end{itemize}

\subsection{Performance}

\begin{itemize}
    \item Llama 70B\_q achieves the highest overall performance using greedy decoding, while Gemma-family models consistently perform the worst across all prompting techniques evaluated.

    \item When used with Guided-CoT, the mid-sized model Mistral 24B demonstrates performance comparable to reasoning models QwQ 32B and DeepSeek-R1-Llama 70B\_q.

    \item Guided-CoT, with greedy decoding, substantially improves antisemitism detection performance across all models compared to Zero-Shot and Zero-Shot-CoT, except Mistral 8B, which shows only marginal improvement.
    
\end{itemize}

\subsection{Reliability}

\begin{itemize}

    \item Zero-Shot prompting results in homogeneous responses, and under Zero-Shot-CoT, notable divergence is seen in reasoning-oriented models.
    Guided-CoT further amplifies latent differences, particularly within smaller models and reasoning models; however, Llama 70B\_q and Mistral 24B are seen very close to each other.

    \item Under the Zero-Shot setting, all models except Llama 70B\_q exhibit significant semantic distance from all the posts classified as antisemitic toward those classified non-antisemitic.
    The absence of this difference in Llama 70B\_q, consistent across Zero-Shot-CoT and Guided-CoT prompts, indicates strong adaptability in following task-specific policies for antisemitism classification.

    \item  Intra-model analyses reveal significant differences in response distributions for antisemitic versus non-antisemitic classifications under Zero-Shot.
    Three models exhibit bimodal or crossing distributions (e.g., Gemma 12B, Llama 8B, Mistral 24B), indicating subsets of posts with complex behaviors for interpretability, requiring deeper qualitative review.
    Strategic interpretability audits guided by these statistical insights are essential for uncovering subtle biases and ensuring robust moderation practices.
    
\end{itemize}

\subsection{Prompting Techniques}

\begin{itemize}
    \item Transitioning from Zero-Shot-{$\alpha$} to Zero-Shot-{$\beta$} (i.e., IHRA definition with contemporary examples) or to Zero-Shot-CoT (adding reasoning instructions along with full IHRA definition) decreases performance for certain models (e.g., Llama 8B, Mistral 24B, and Llama 70B\_q), indicating that simply providing more context or prompting model to reason does not ensure improvements.

    \item Guided-CoT consistently yields the highest F1 gains regardless of the decoding strategy chosen, except Mistral 8B with a greedy decode.
    Our ablation study suggests a potential reason for this: removing one thought may have improved Mistral 8B's performance by 0.04 on F1.

    \item Prompting techniques substantially influence how closely each model’s explanations align, on average, with those of other models.
    However, no consistent pattern emerges across different prompting methods.
    Llama 70B\_q consistently generates the most cohesive responses close to other models, regardless of the prompting technique.    
    
\end{itemize}

\begin{figure*}[t]
\centering
\includegraphics[width=0.98\linewidth]{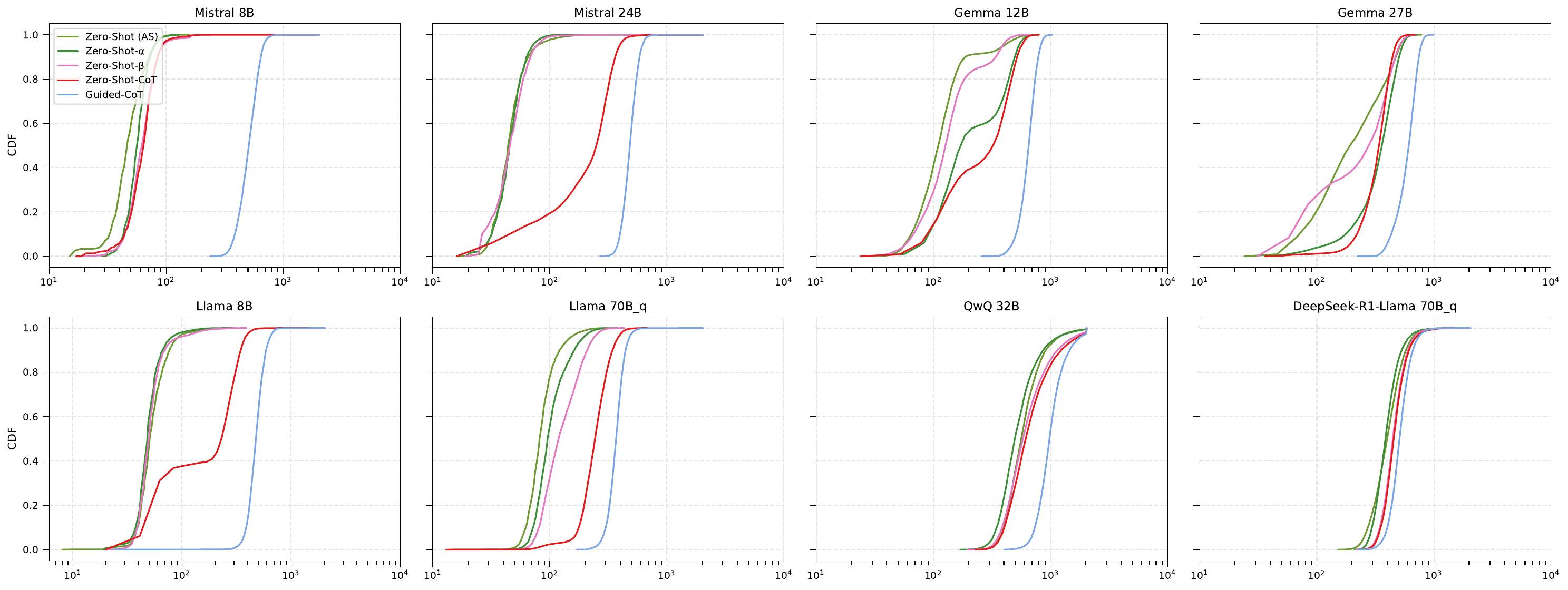}
\caption{CDF of total number of generated tokens for all posts across eight models and all prompting variants. NB: X-axis is log-transformed (i.e., $10^x$).
Among non-reasoning models, Gemma-family models generate the longest responses.
QwQ 32B generates the longest responses out of the reasoning models.
Guided-CoT generates longer responses than other prompting techniques studied.}
\label{fig:model-wise-output-tokens}
\end{figure*}

\section{Models}
\label{apx:models}

This study evaluates various state-of-the-art large language models for their capability to detect antisemitism.
We select eight models of different sizes and from different families:

\begin{itemize}

  \item Mistral 2410~\footnote{Mistral 8B:~\url{https://mistral.ai/news/ministraux}} (8B) and Mistral 2501~\footnote{Mistral 24B:~\url{https://mistral.ai/news/mistral-small-3}} (24B) from Mistral AI,
  \item Gemma 3~\footnote{Gemma 12B \& 27B:~\url{https://arxiv.org/abs/2503.19786}} (12B and 27B) from Google,
  \item Llama 3.1~\footnote{Llama 8B \& 70B:~\url{https://arxiv.org/abs/2407.21783}} (8B and 70B) from Meta,
  \item DS-R1-Llama~\footnote{DS-R1-Llama 70B:~\url{https://arxiv.org/abs/2501.12948}} (70B) from DeepSeek, a multilingual distilled model (based on Llama 3.3 70B) explicitly incorporating reasoning abilities,
  \item QwQ~\footnote{QwQ 32B:~\url{https://qwenlm.github.io/blog/qwq-32b/}} (32B) from Qwen, another multilingual reasoning-capable model.

\end{itemize}

Note that we utilize quantized versions for both 70B variants, Llama 3.1 70B Instruct quantized~\footnote{\url{https://huggingface.co/RedHatAI/Meta-Llama-3.1-70B-Instruct-quantized.w4a16}} and DS-R1-Llama~\footnote{\url{https://huggingface.co/RedHatAI/DeepSeek-R1-Distill-Llama-8B-quantized.w4a16}}, improving efficiency and enabling evaluation of all models on a single GPU due to resource-constrained settings.
All the models we evaluate in this study are intruct (chat) models.
Table \ref{apx:model-attributes} summarizes the attributes of the models we evaluate in this study.

\subsection{Model's understanding of Antisemitism}
\label{apx:antisemitism-definition}

We prompt each model with the question, \textbf{\textit{``What is the definition of antisemitism?''}} to assess their understanding and examine alignment with the widely-used IHRA definition in this study.
Most models explicitly reference the IHRA definition in their responses, except for Llama 8B, Mistral 24B, and DeepSeek-R1.
\begin{table*}[t]
    \centering
    \begin{tabular}{l c c c c c}
    \toprule
    \textbf{Model} & \textbf{ZS-(AS)} & \textbf{ZS-{\boldmath$\alpha$}} & \textbf{ZS-{\boldmath$\beta$}} & \textbf{ZS-CoT} & \textbf{Guided-CoT} \\
    \midrule
    
    Mistral 8B & 49 & 57 & 64 & 65 & 513 \\
    Mistral 24B & 48 & 47 & 48 & 228 & 489 \\
    Gemma 12B  & 136 & 257 & 156 & 301 & 650 \\
    Gemma 27B & 238 & 350 & 263 & 335 & 600 \\
    Llama 8B  & 56 & 51 & 54 & 196 & 483  \\
    Llama 70B\_q & 89 & 107 & 134 & 251 & 372 \\
    
    \midrule
    
    QwQ 32B & 628 & 573 & 694 & 734 & 1,048 \\
    DS-R1-Llama 70B\_q & 416 & 405 & 464 & 474 & 523  \\
    
    \bottomrule
    \end{tabular}
    \caption{Average number of generated tokens per prompt for a single social media post. Reasoning models produce more tokens due to built-in thinking capability before generating responses.}
    \label{tab:token-per-model-prompt}
\end{table*} 

\section{Differences in Generated Responses}
\label{apx:diff-in-generated-responses}

\subsection{Length} 

Fig.~\ref{fig:model-wise-output-tokens} illustrates the cumulative distribution functions (CDF) for generated tokens across prompting techniques.
Using greedy decoding for 11,311 posts, we generate approximately 143 million tokens across eight models and five prompting variants.
Table~\ref{tab:token-per-model-prompt} provides mean response lengths (output tokens) generated by different models across these techniques.

The change from Zero-Shot prompts to more sophisticated prompts (i.e., Guided-CoT) generally increases response length, especially among non-reasoning models.
Notably, the Gemma family consistently generates longer responses than other non-reasoning models for the same prompting variants.
Additionally, Mistral 8B fails to generate a reasoning chain in the ZS-CoT setting despite including the explicit instruction ``Let's think step by step,'' and we found that forcing the model to reason required presenting this phrase independently without any additional text surrounding this phrase.
Guided-CoT consistently generates the longest responses across all models studied.
On average, Guided-CoT generates approximately twice the number of tokens compared to ZS-CoT. 
Reasoning models generate longer responses across all prompting strategies due to inclusion of thinking tokens.
Specifically, QwQ 32B generates very long responses, highlighting a substantial generation of thinking tokens before providing an answer.
QwQ 32B generates about 1.5 times more tokens with Guided-CoT than ZS-CoT, whereas DeepSeek-R1-Llama 70B\_q shows only a marginal increase.

\begin{figure}[t]
\centering
\begin{tcolorbox}[width=0.8\columnwidth, colframe=red!50!black, colback=red!5!white, title=Refusal Response (Llama 8B with ZS-{$\beta$}), fonttitle=\scriptsize, fontupper=\footnotesize]
\small
I cannot create content that could be used to promote or glorify antisemitism. Is there anything else I can help you with?
\end{tcolorbox}
\caption{}
\label{fig:refusal-example}
\end{figure}

\subsection{Generation} 

We categorize model responses into three groups, as discussed in \S~\ref{sec:diff-in-responses-main}.
Further examination reveals distinct response patterns across models, indicating varying behaviors likely arising from their specific training or post-alignment methods.
Gemma-family models primarily generate responses categorized as ``indeterminate,'' offering uncertain or context-dependent classifications.
Interestingly, Gemma-family models show reduced invalid responses with CoT-like prompting, Zero-Shot-CoT, and Guided-CoT.
We hypothesize that CoT-style prompts may compel these models to recognize ambiguity, leading them to provide cautious or non-committal responses when the context is insufficient.
In contrast, Llama 8B often explicitly refuses to generate responses, interpreting the given task as generating antisemitic posts rather than classifying them (see Fig.~\ref{fig:refusal-example}).
QwQ 32B frequently exceeds the 2048-token response limit due to getting stuck in repetitive token generation under greedy decoding and this means we don't get an answer.

\begin{figure*}[t]
  \centering
    \begin{subfigure}[b]{0.3\textwidth}
      \includegraphics[width=\textwidth]{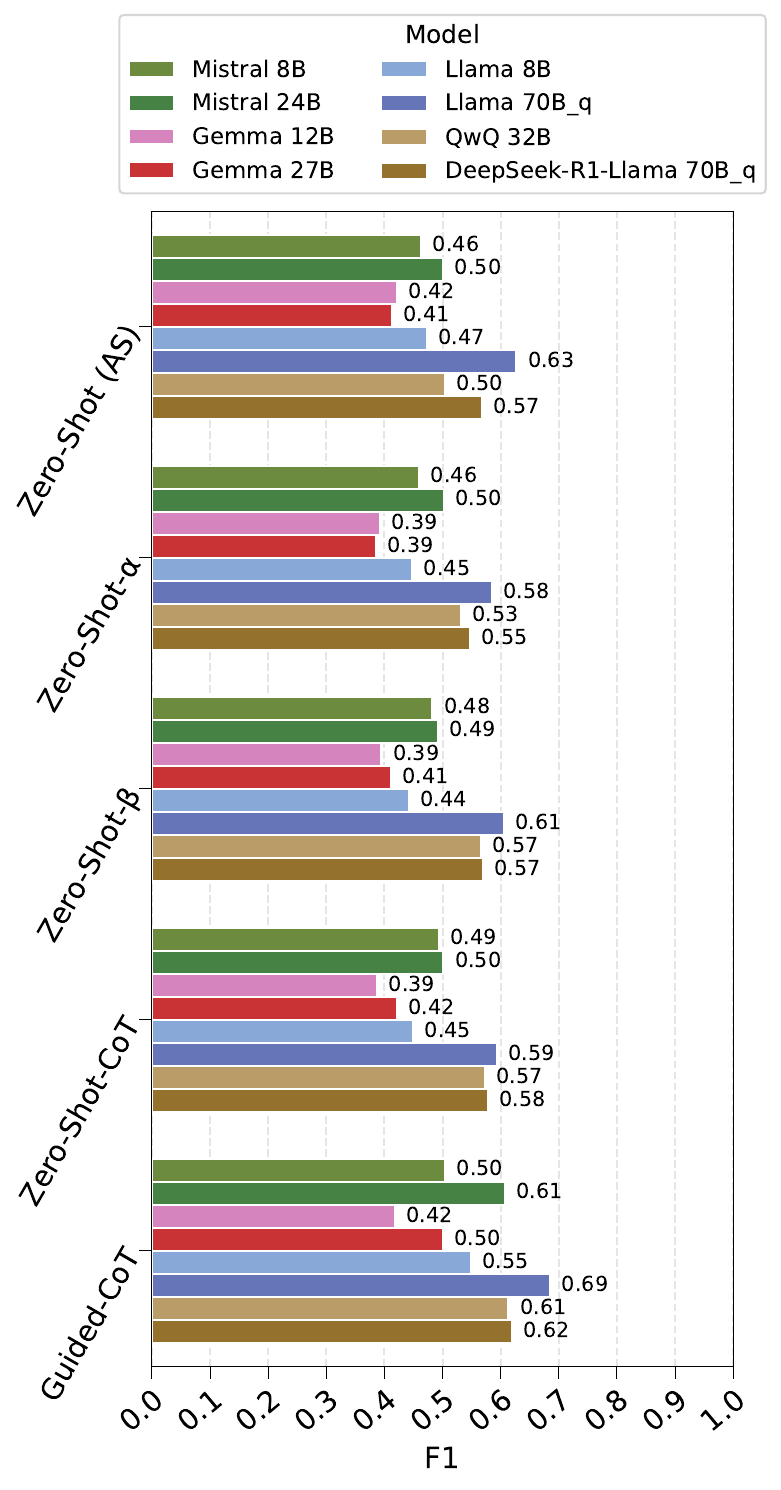}
      \caption{}
      \label{fig:}
    \end{subfigure}
    \begin{subfigure}[b]{0.3\textwidth}
      \includegraphics[width=\textwidth]{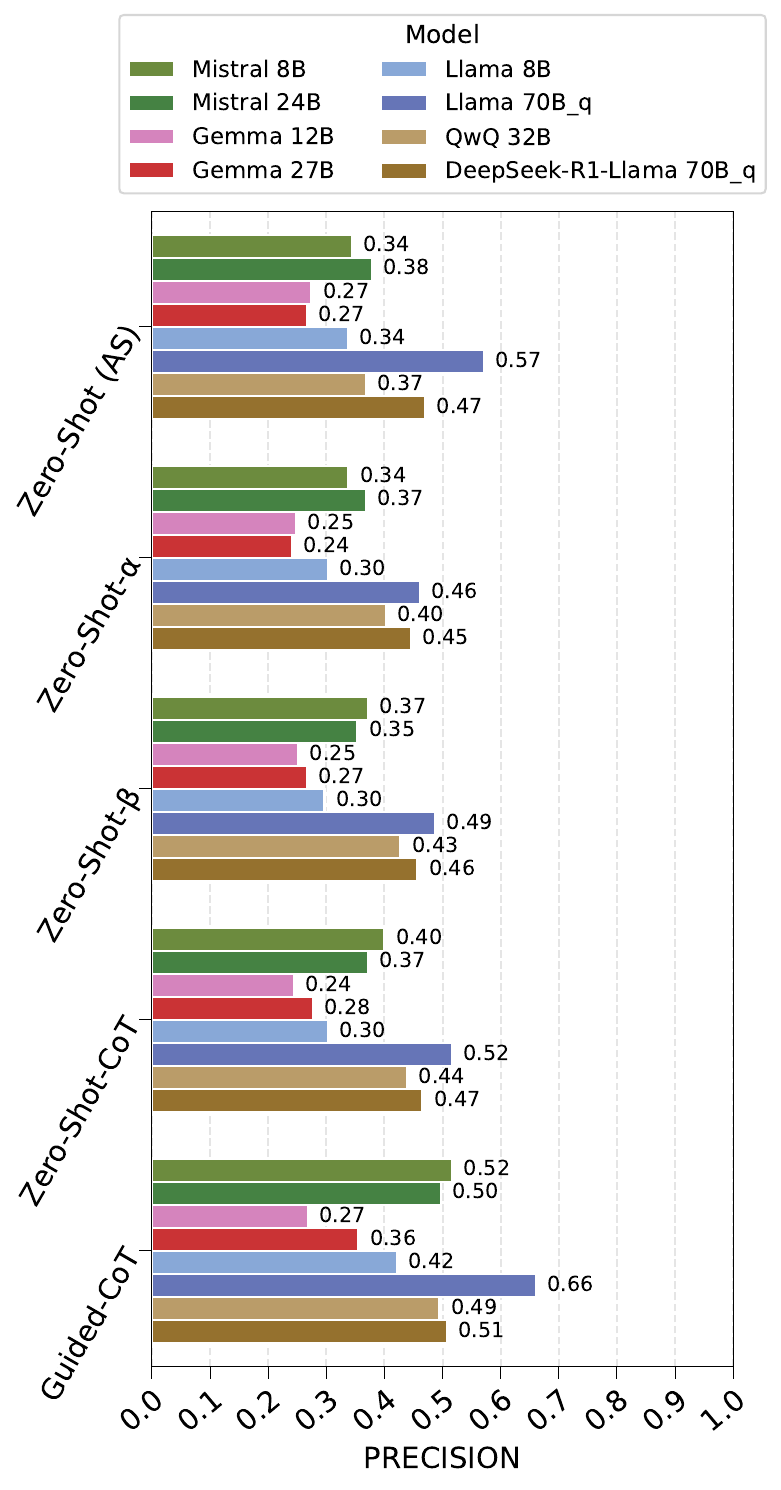}
      \caption{}
      \label{fig:}
    \end{subfigure}
    \begin{subfigure}[b]{0.3\textwidth}
      \includegraphics[width=\textwidth]{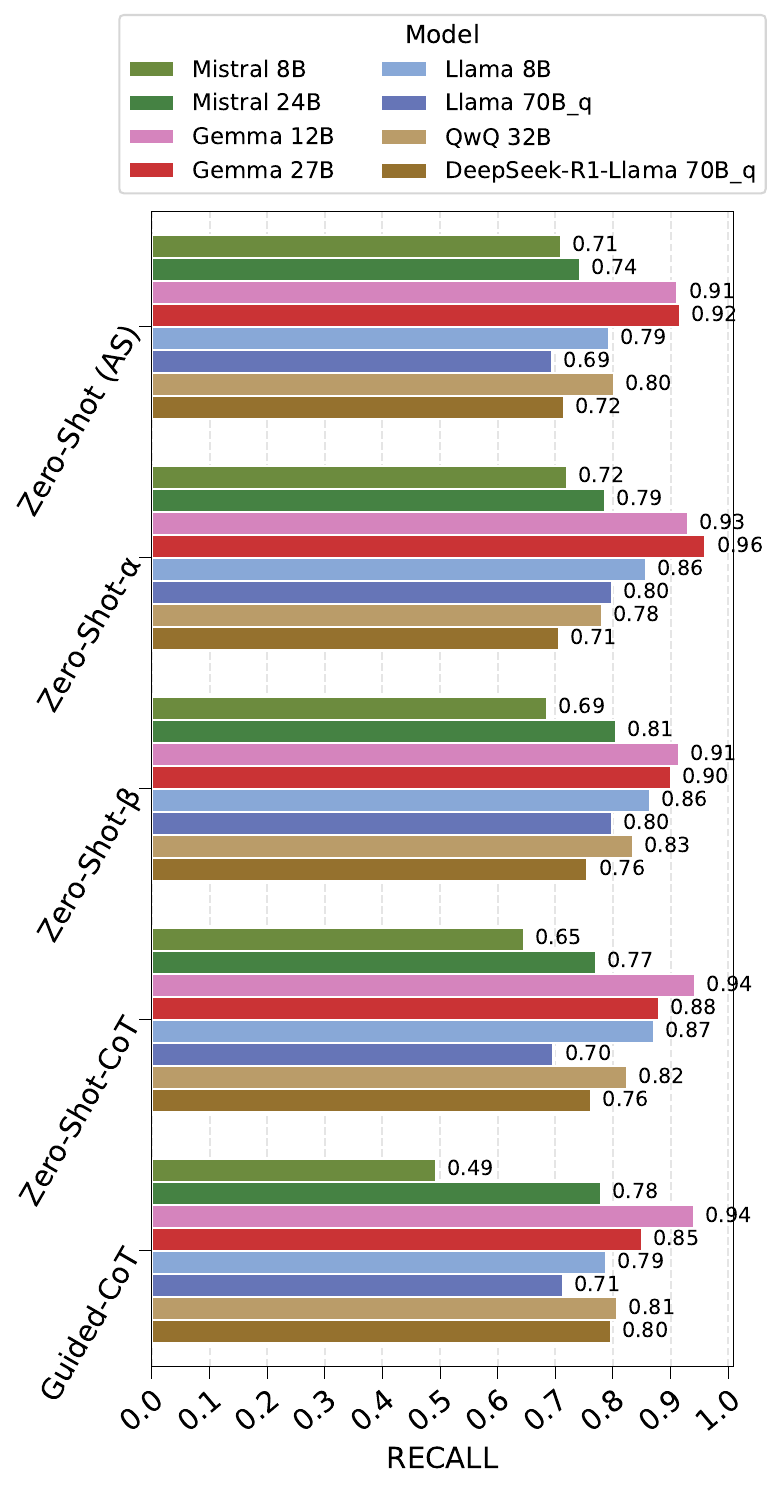}
      \caption{}
      \label{fig:recall}
    \end{subfigure}
  \caption{\textbf{Setup:} Greedy decoding, 8,555 posts (7,031 non-antisemitic, 1,524 antisemitic).
  \textbf{Metrics:} F1-score, precision, and recall (antisemitic class).
  Guided-CoT consistently outperforms other prompts across all models.
  NB: We exclude invalid responses from the analysis.}
  \label{fig:common-points-models-performance-greedy}
\end{figure*}

\begin{figure*}[t]
    \centering
      \begin{subfigure}[b]{0.3\textwidth}
        \includegraphics[width=\textwidth]{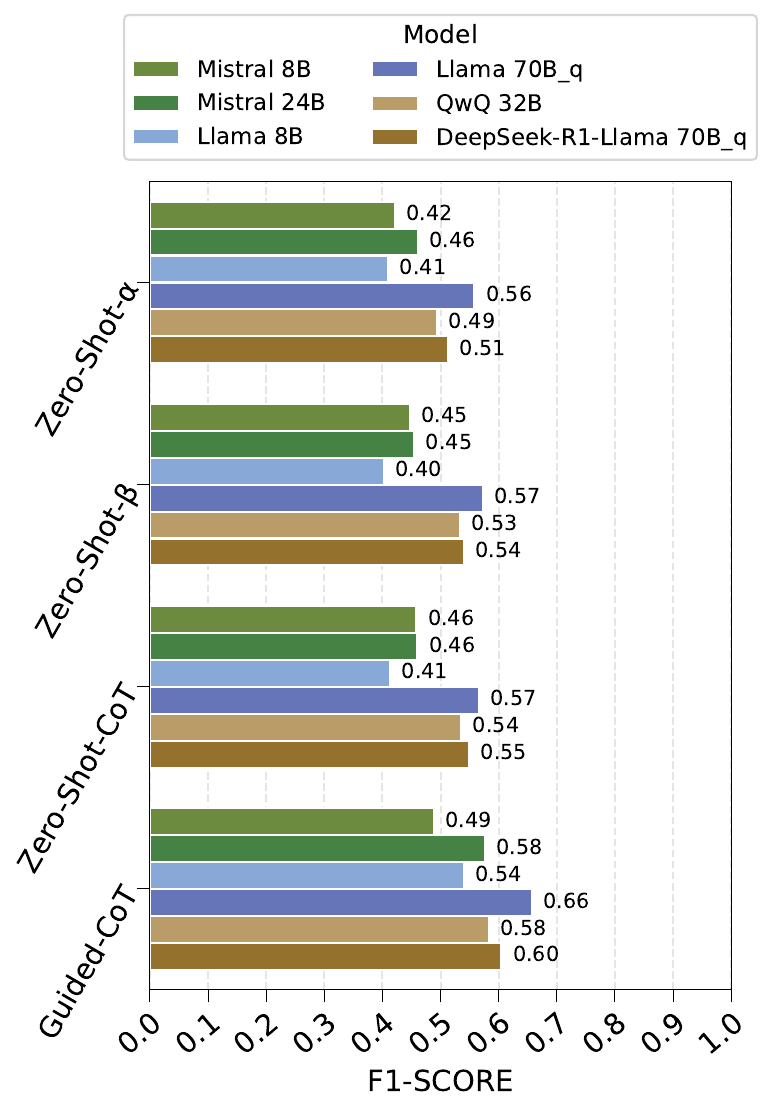}
        \caption{}
        \label{fig:}
      \end{subfigure}
      \begin{subfigure}[b]{0.3\textwidth}
        \includegraphics[width=\textwidth]{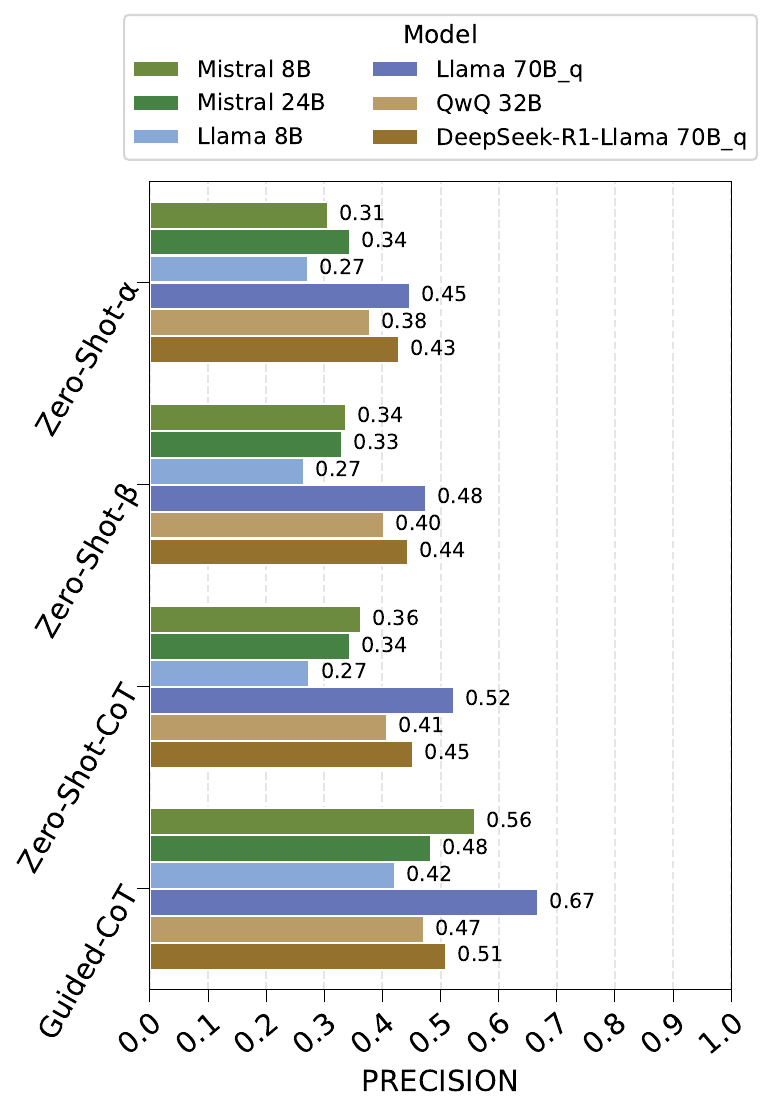}
        \caption{}
        \label{fig:}
      \end{subfigure}
      \begin{subfigure}[b]{0.3\textwidth}
        \includegraphics[width=\textwidth]{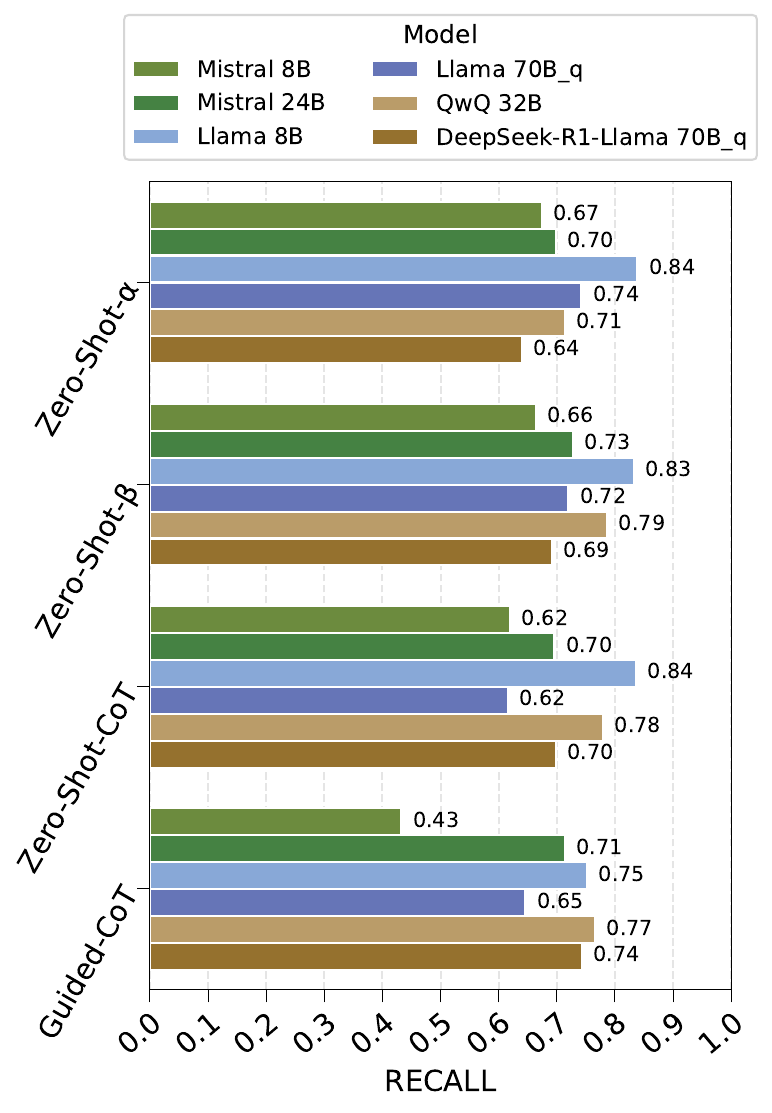}
        \caption{}
        \label{fig:}
      \end{subfigure}
    \caption{\textbf{(a, b \& c)} \textbf{Setup:} Self-consistency decoding, 11,311 posts (7,031 non-antisemitic, 1,524 antisemitic), 30 runs per post (temperature=0.6, top\_p=0.9), final prediction by majority vote.
    \textbf{Metric:} F1-score, precision, recall (antisemitic class).
    Guided-CoT consistently outperforms other prompting methods.
    Red indicates decreased performance; green indicates improvement.
    ZS-{$\beta$} decreases performance for Mistral 24B and Llama 8B; ZS-CoT marginally decreases performance for Mistral 24B only.
    }
    \label{fig:model-performance-SC}
  \end{figure*}

\section{Deterministic \& Non-deterministic Evaluation}
\label{apx:deter-non-deter-eval}

We compare prompting techniques for deterministic and non-deterministic evaluations for precision and recall in Fig.~\ref{fig:common-points-models-performance-greedy} and Fig.~\ref{fig:model-performance-SC}.
Note that Self-consistency metrics are calculated over a complete dataset (11,315 posts).
By running multiple inferences for a single social media post using non-deterministic hyper parameters, we get at least one valid response for a given post.

\begin{table*}[t]
    \centering
    \begin{tabular}{cccc}
    
    \hline
     & \textbf{Llama 3.1 70B-q} & \textbf{Llama 3.1 70B-q} & \textbf{fine-tuned GPT-3.5} \\ 
    
    \#samples & 6,940 & 6,941 & 6,941 \\ 
    
    hyperparams & Greedy & Self-consistency & - \\ 
    \hline
    
    not-AS 
      & 0.95 / 0.91 / \textbf{0.93}
      & 0.95 / 0.92 / \textbf{0.93}
      & 0.95 / 0.89 / \textbf{0.92} \\
    
    AS 
      & 0.67 / 0.78 / \textbf{0.72}
      & 0.68 / 0.78 / \textbf{0.73}
      & 0.62 / 0.80 / \textbf{0.70} \\
    
    Overall 
      & 0.81 / 0.85 / \textbf{0.83}
      & 0.82 / 0.85 / \textbf{0.83}
      & 0.79 / 0.85 / \textbf{0.81} \\ 
    \hline 
    
    \hline
    \end{tabular}
    \caption{Comparing our best-performing prompting technique (i.e., Guided-CoT) with Greedy/self-consistency decoding to fine-tuned GPT-3.5~\citep{becker2024antisemitism}.
    The numbers represent precision, recall, and F1 (i.e., P/R/F1).
    Guided-CoT with greedy or SC decoding outperforms the fine-tuned GPT-3.5 model.}
    \label{tab:all-guided-cot-stats}
  \end{table*}

\section{Guided-CoT Ablation Setup}
\label{apx:guided-cot-more}

The ablation experiments are defined as follows:

\begin{itemize}

  \item \textbf{A1:} Ablation 1 excludes breaking the social media post into chunks.

  \item \textbf{A2} excludes guiding the model to check for sarcasm, criticism, quoted statements, news, coded language, and ambiguous phrasing in the author's post.

  \item \textbf{A3} excludes identifying whether the author is calling to promote antisemitism or using their narrative to spread awareness about antisemitism.

  \item \textbf{A4} excludes the reminder to the model that classification must align with the IHRA definition and contemporary examples.

  \item \textbf{A5} excludes a guide to use the overall analysis for judging the author's stance toward the Jewish community.

\end{itemize}

We exclude one thought at a time (out of 5 thoughts), for example, prompt ``A1'' excludes thought A1 while retaining thoughts A2, A3, A4, and A5.

\section{Kolmogorov--Smirnov (KS) Significance Tests Setup}
\label{apx:significance-details}

We conduct a significance test to determine if the variations observed between $D^{+}$ and $D^{-}$ distributions are statistically significant.

\begin{align*}
D^{+} &= \{\,d_i^{(k)} : i \in \{\text{posts classified \textit{antisemitic}}\}, \\
      &\hspace{1em} k=1500 \} \\
D^{-} &= \{\,d_i^{(k)} : i \in \{\text{posts classified \textit{non-AS}}\}, \\
      &\hspace{1em} k=1500 \} \\
d_i^{(k)} &= \frac{1}{k}\sum_{j \in \mathcal{N}_k(i)}
            \operatorname{\mathcal{D}_c}(\mathbf{e}_i,\mathbf{e}_j)
\end{align*}

Let \(F^{+}(x)=P\!\bigl(D^{+}\le x\bigr)\) and \(F^{-}(x)=P\!\bigl(D^{-}\le x\bigr)\) denote the ECDFs of the fixed-\(k\) cohesion scores for posts classified as antisemitic and non-antisemitic, respectively.
We perform three variants of the two-sample KS hypothesis tests:

\subsubsection*{\textbf{Two-sided test:}}

\begin{align*}
  H_{0}: &\quad F^{+}(x) = F^{-}(x) \quad \text{for all } x \\
  H_{a}: &\quad F^{+}(x) \neq F^{-}(x) \quad \text{for some } x .
\end{align*}

\textit{Intuition:} Do the two ECDFs differ \emph{anywhere}?  
A significant result means the shapes are not identical, there is at least one range of cohesion scores where the two groups diverge.

\subsubsection*{\textbf{One-sided (greater) test:}}

\begin{align*}
  H_{0}: &\quad F^{+}(x) \le F^{-}(x) \quad \text{for all } x \\
  H_{a}: &\quad F^{+}(x) > F^{-}(x) \quad \text{for some } x .
\end{align*}
\textit{Intuition:} Is the antisemitic curve shifted \emph{left} (toward lower scores)?  
Rejecting \(H_{0}\) implies antisemitic explanations are, on average, \textbf{more cohesive} (semantically closer, meaning distance is less) than non-antisemitic ones.

\subsubsection*{\textbf{One-sided (less) test:}}

\begin{align*}
  H_{0}: &\quad F^{+}(x) \ge F^{-}(x) \quad \text{for all } x \\
  H_{a}: &\quad F^{+}(x) < F^{-}(x) \quad \text{for some } x .
\end{align*}
\textit{Intuition:} Is the antisemitic curve shifted \emph{right} (toward higher scores)?  
Rejecting \(H_{0}\) indicates antisemitic explanations are, on average, \textbf{less cohesive} (at distant) than their non-antisemitic counterparts.

\textbf{Note:} If the two-sided test is significant but \emph{only one} one-sided test is significant, the difference is monotonic: antisemitic responses are either consistently closer or consistently distant.
If \emph{both} one-sided tests are significant, the ECDFs must cross.  
Antisemitic responses are closer in one part of the distribution but distant in another, indicating a complex and a non-monotonic (bimodal or skewed) pattern that merits qualitative inspection.
  
\begin{table*}[t]
  \small
  \centering
  \begin{tabular}{l|l|l}
  \toprule
  \textbf{ZS-{$\beta$}} & \textbf{ZS-CoT} & \textbf{Guided-CoT} \\
  \midrule
  
  Llama 70B\_q (0.11) & Llama 70B\_q (0.13) & Llama 70B\_q (0.17) \\
  Mistral 8B (0.11) & Llama 8B (0.14) & Mistral 24B (0.18) \\
  Mistral 24B (0.12) & Mistral 8B (0.14) & QwQ 32B (0.22) \\
  Llama 8B (0.13) & Gemma 27B (0.15) & Gemma 12B (0.23) \\
  Gemma 27B (0.15) & Mistral 24B (0.15) & DS-R1-Llama 70B\_q (0.27) \\
  QwQ 32B (0.15) & DS-R1-Llama 70B\_q (0.18) & Gemma 27B (0.31) \\
  Gemma 12B (0.16) & QwQ 32B (0.19) & Llama 8B (0.37) \\
  DS-R1-Llama 70B\_q (0.20) & Gemma 12B (0.29) & Mistral 8B (0.38) \\
  
  \bottomrule
  \end{tabular}
  \caption{Average median distance per model to all other models, sorted by lowest to highest.}
  \label{tab:sorted-model-distance}
\end{table*}

\begin{figure*}[t]
\centering
\includegraphics[width=0.98\linewidth]{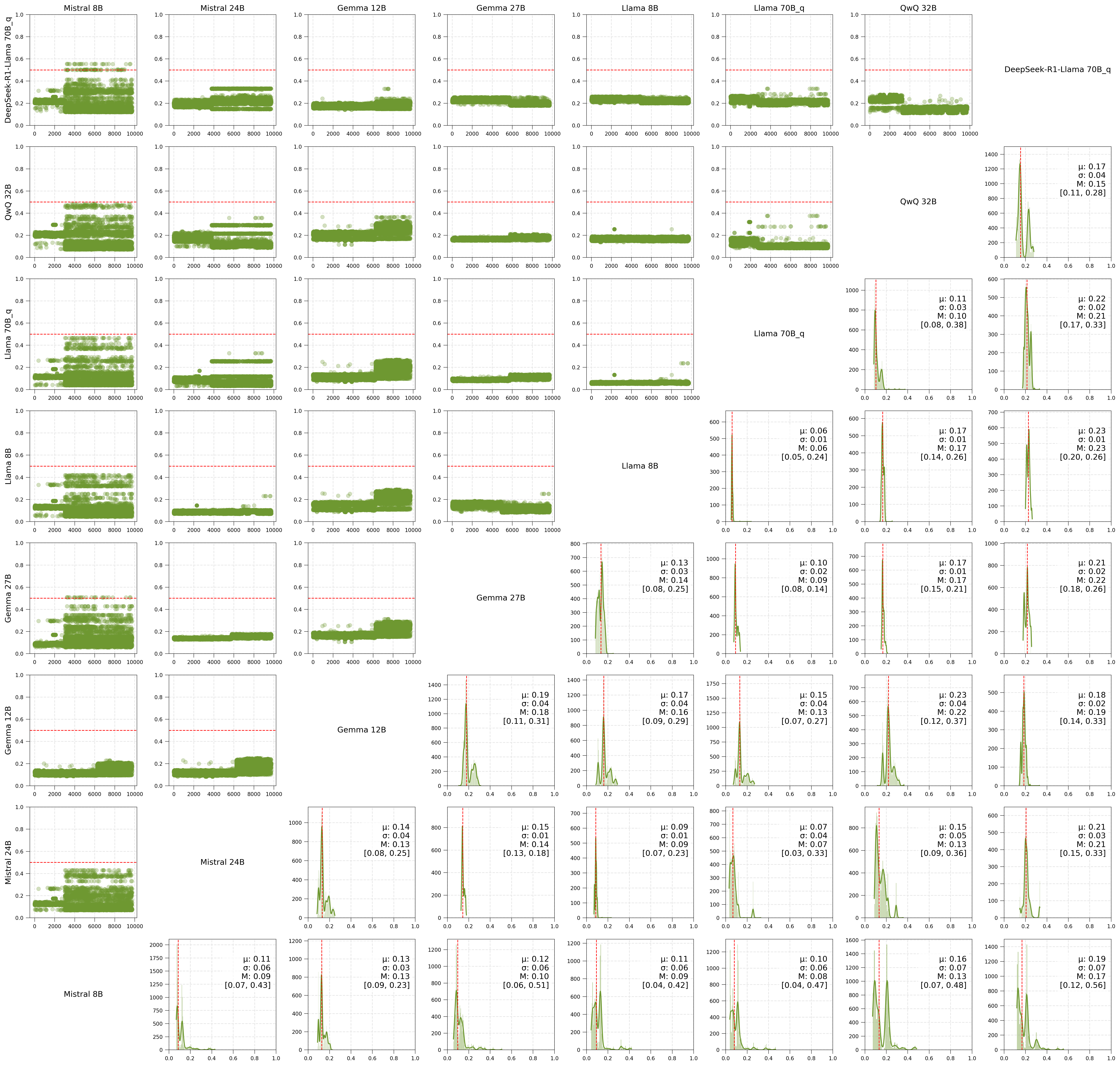}
\caption{Distribution (Eq.~\ref{eq:cos_dist}) of normalized cosine distance between the responses generated by models. The upper left half shows average differences between model pairs, and the bottom half displays histograms with KDE curves. NB: Reasoning models' responses are embeded without thinking tokens. \textbf{Prompting:} ZS-{$\beta$}.}
\label{fig:zs-beta-facet}
\end{figure*}

\begin{figure*}[t]
\centering
\includegraphics[width=0.98\linewidth]{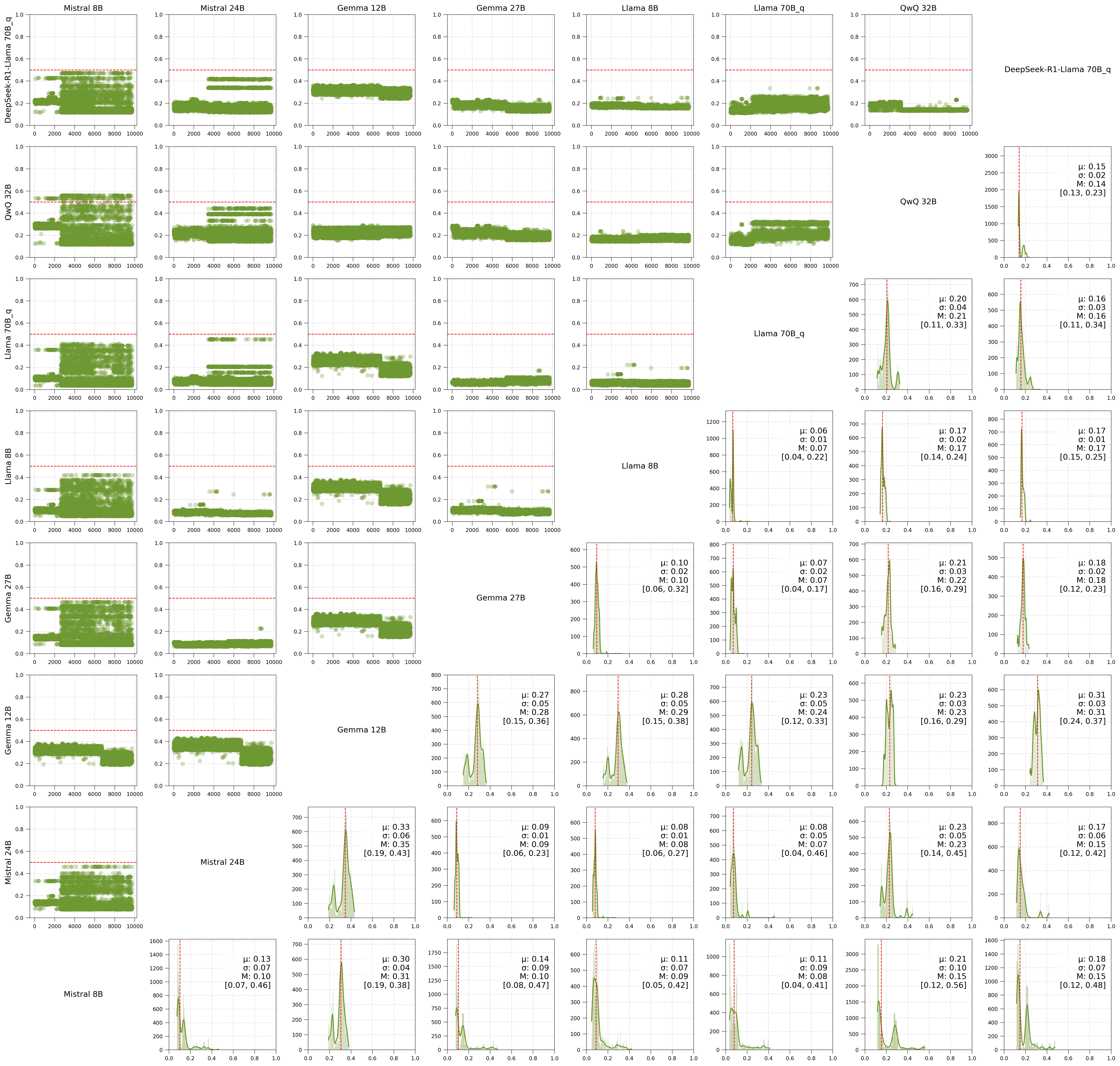}
\caption{Distribution (Eq.~\ref{eq:cos_dist}) of normalized cosine distance between the responses generated by models. The upper left half shows average differences between model pairs, and the bottom half displays histograms with KDE curves. NB: Reasoning models' responses are embeded without thinking tokens. \textbf{Prompting:} ZS-CoT.}
\label{fig:zs-cot-facet}
\end{figure*}

\begin{figure*}[t]
\centering
\includegraphics[width=0.98\linewidth]{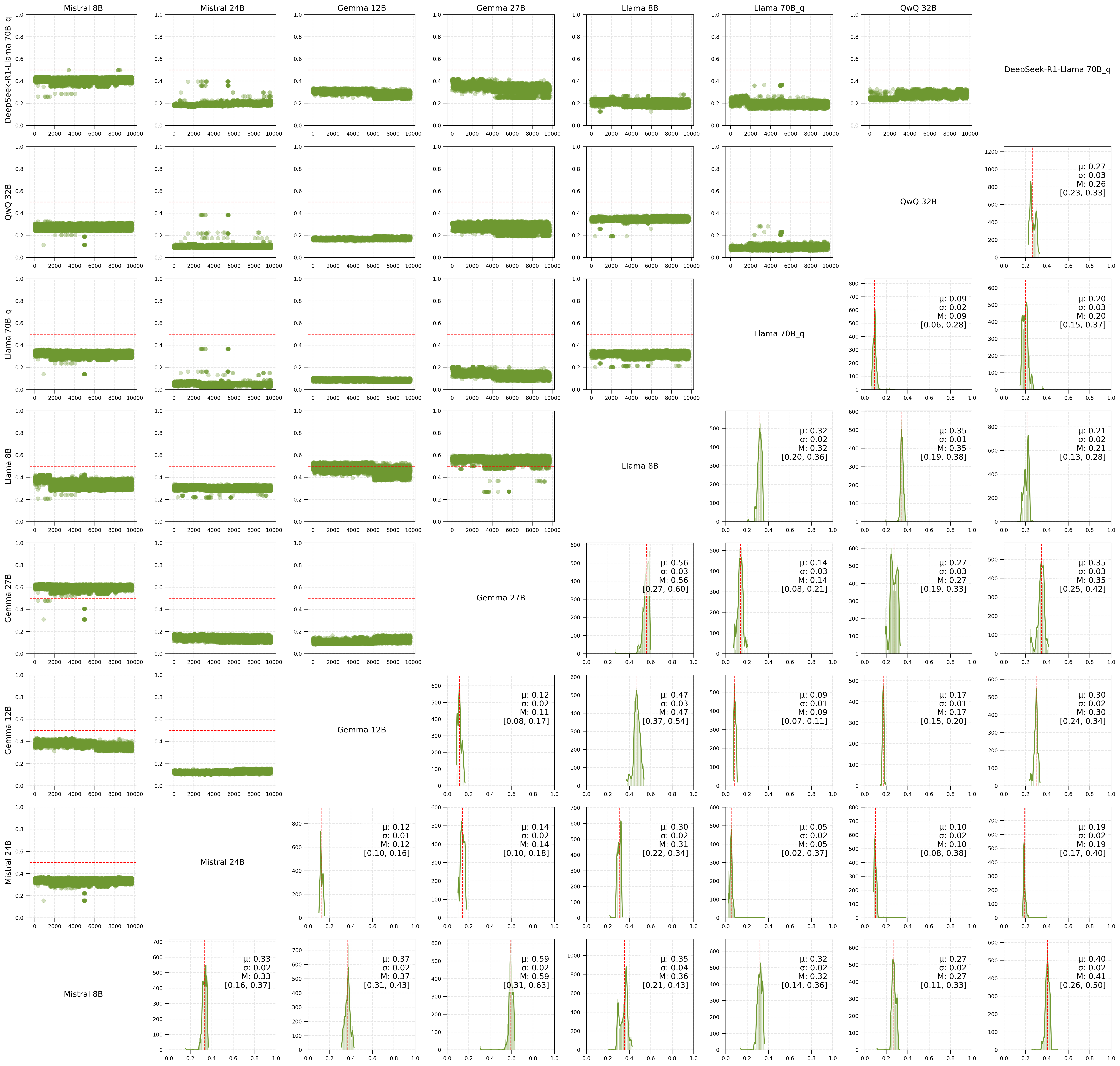}
\caption{Distribution (Eq.~\ref{eq:cos_dist}) of normalized cosine distance between the responses generated by models. The upper left half shows average differences between model pairs, and the bottom half displays histograms with KDE curves. NB: Reasoning models' responses are embeded without thinking tokens. \textbf{Prompting:} Guided-CoT.}
\label{fig:guided-cot-facet}
\end{figure*}

\section{Annotation Codebook}
\label{apx:codebook}

\textbf{Sampling:} From 8,555 valid responses across all eight models, we find that 260 posts were misclassified by all eight models under Guided-CoT prompting (performance evaluation discussed in \S~\ref{sec:deterministic-eval}).

\textbf{Annotation:} Each annotator was given only a Twitter post and options to choose one of the appropriate categories for each post.
The annotations were conducted using the online version of Google Docs.
The codebook for annotating posts in the misclassification analysis:

\begin{itemize}

\item \textbf{Tone/Language/Stereotypes:} Posts that target the Jews individually or as a group using a tone, offensive language, or stereotypes.

\item \textbf{Quotes/Reporting (Q/NR):} Posts that quote knowledge of information of someone else (not OP's opinion) or report another source (e.g., news articles, link to other source) with clear attribution (quotation marks, named speaker/source, cited outlet).

\item \textbf{Opinion:} Posts that express personal views or sentiments rather than purely objective facts.

\item \textbf{Religious:} Posts that discuss religious beliefs.

\item \textbf{Sarcasm:} Posts that use ironic or nonliteral language to convey the opposite of what is literally written by the OP.

\item \textbf{Contextual Mismatch:} Posts with typos or off-topic (i.e., not antisemitism related entities) context.

\item \textbf{Other:} Posts that do not fit any of the above categories.

\end{itemize}

\begin{table*}[t]
\small
\centering
\begin{tabular}{p{13cm}}
\toprule
\textbf{Llama 70B\_q} \\
\midrule

\textbf{Twitter Post:}
@CathyHigginsWY @VP @POTUS Whoaaaaaaaaaa. This makes me question my whole belief system. Next you're gonna tell me that Soros doesn't control all Jews and that rural folk aren't subsidized by the devil worshipping city fags who eat curry and smoke jazz cigarettes \\

\textbf{Ground Truth:} Non Antisemitic
\\

\hline

\textbf{Zero-Shot-$\beta$:}
\newline\newline
This Twitter post contains several elements that align with the IHRA definition of antisemitism. Specifically, it mentions George Soros, a Jewish billionaire often targeted by antisemitic conspiracy theories, and implies that he controls ``all Jews.'' This is a classic example of a mendacious and stereotypical allegation about Jews as a collective, which is listed as an example of antisemitism in the IHRA definition.\newline
Additionally, the post employs sinister stereotypes and negative character traits, such as implying that Jews are part of a group that is ``devil worshipping'' and engaging in other derogatory behaviors. This kind of language is also consistent with the IHRA definition of antisemitism.
\newline\newline
Summary: The post contains antisemitic language and conspiracy theories.  
\newline
Antisemitic: Yes 
\newline
\\

\hline

\textbf{Zero-Shot-CoT:}
\newline\newline
To classify the Twitter post, let's analyze it step by step according to the IHRA definition of antisemitism.\newline 1. The post mentions ``Soros,'' likely referring to George Soros, a Jewish billionaire often targeted by antisemitic conspiracy theories.\newline 2. The phrase ``Soros doesn't control all Jews'' implies that Soros has control over some or many Jews, which is a classic antisemitic trope about Jewish people being controlled by a single individual or group.\newline 3. This statement is an example of making ``mendacious, dehumanizing, demonizing, or stereotypical allegations about Jews as such or the power of Jews as collective'' (IHRA definition).\newline Based on the analysis, the post contains antisemitic content.\newline \newline Summary: The Twitter post contains antisemitic content by perpetuating a conspiracy theory about Jewish people being controlled by George Soros.
\newline
Antisemitic: Yes
\newline
\\

\hline

\textbf{Guided-CoT:}
\newline
Let's break down the Twitter post step by step:\newline \newline 1. The post is broken into several chunks:\newline    - The initial expression of surprise and questioning of the author's belief system.\newline    - The mention of Soros, implying a connection to Jews.\newline    - The statement about rural folk being subsidized by "devil-worshipping city fags who eat curry and smoke jazz cigarettes," which appears to be a derogatory and unrelated remark.\newline \newline 2. The post appears to be a mix of sarcasm and coded language. The author seems to be using irony and hyperbole to express their surprise and skepticism. However, the mention of Soros and the connection to Jews is a common antisemitic trope.\newline \newline 3. The post does not appear to be calling out to promote antisemitism or using statements to spread awareness about antisemitism. Instead, it seems to be using coded language to perpetuate antisemitic stereotypes.\newline \newline 4. The post potentially aligns with the IHRA definition of antisemitism, specifically the example of "Making mendacious, dehumanizing, demonizing, or stereotypical allegations about Jews as such or the power of Jews as collective — such as, especially but not exclusively, the myth about a world Jewish conspiracy or of Jews controlling the media, economy, government or other societal institutions." The mention of Soros controlling Jews is a common antisemitic conspiracy theory.\newline \newline 5. Based on the analysis, the post's stance is not favorable to the Jewish community, as it perpetuates antisemitic stereotypes and conspiracy theories.\newline \newline Summary: The post contains antisemitic language and perpetuates conspiracy theories about Jews.
\newline
\textbf{Antisemitic:} Yes
\newline
\\

\bottomrule
\end{tabular}
\caption{[False Positive (Tone/Language/Stereotypes), Guided-CoT] Sample response across prompting techniques for Llama 70B\_q.}
\label{}
\end{table*}

\begin{table*}[t]
\small
\centering
\begin{tabular}{p{13cm}}
\toprule
\textbf{Llama 70B\_q} \\
\midrule

\textbf{Twitter Post:}
RT @Purbita9: \#AdiZaMountThisAward Aditi Rathore is daring she directly kikes adiza VMS \\

\textbf{Ground Truth:} Non Antisemitic
\\

\hline

\textbf{Zero-Shot-$\beta$:}
\newline\newline
The Twitter post contains the term ``kikes,'' which is a derogatory and offensive term used to refer to Jews. This term is a classic example of antisemitic language and is often used to express hatred or contempt towards Jews.\newline \newline Summary: The Twitter post contains antisemitic language.
\newline
Antisemitic: Yes 
\newline
\\

\hline

\textbf{Zero-Shot-CoT:}
\newline
To classify the Twitter post, let's break it down step by step:\newline \newline 1. The post contains the word ``kikes,'' which is a derogatory and offensive term used to refer to Jews.\newline 2. The IHRA definition states that antisemitism can be expressed in speech, writing, visual forms, and action, and employs sinister stereotypes and negative character traits.\newline 3. The use of the term ``kikes'' is a classic example of antisemitic language, as it is a pejorative term that has been used to demean and dehumanize Jews throughout history.\newline \newline Based on the IHRA definition, the Twitter post can be classified as antisemitic.\newline \newline Summary: The Twitter post contains antisemitic language and employs a derogatory term to refer to Jews.
\newline
Antisemitic: Yes
\newline
\\

\hline

\textbf{Guided-CoT:}
\newline
Let's analyze the Twitter post step by step:\newline \newline 1. Breaking the post into chunks:\newline    - The post starts with ``RT @Purbita9'', indicating that it's a retweet of someone else's content.\newline    - The hashtag ``\#AdiZaMountThisAward'' seems to be related to a specific topic or event, possibly an award or a competition.\newline    - The text ``Aditi Rathore is daring she directly kikes adiza VMS'' is the main content of the post.\newline \newline 2. Identifying the type of post:\newline    - The post appears to be a statement or an opinion, rather than sarcasm, criticism, news, or reporting.\newline    - The language used is informal and seems to be from a personal perspective.\newline \newline 3. Checking the intent:\newline    - The post does not seem to be calling out to promote antisemitism or using statements to spread awareness about antisemitism.\newline    - However, the use of the word ``kikes'' is a derogatory term for Jews, which raises concerns.\newline \newline 4. Assessing alignment with IHRA definition:\newline    - The post does not directly target the Jewish community, Israel, or Jewish institutions.\newline    - However, the use of the word ``kikes'' is a classic antisemitic trope, which is listed as an example of antisemitism in the IHRA definition.\newline \newline 5. Deciding on the stance:\newline    - The post's use of a derogatory term for Jews suggests that the author's stance is not favorable to the Jewish community.\newline \newline Summary: The post contains a derogatory term for Jews, which raises concerns about antisemitism.
\newline
\textbf{Antisemitic:} Yes
\newline
\\

\bottomrule
\end{tabular}
\caption{[False Positive (Contextual Mismatch), Guided-CoT] Sample response across prompting techniques for Llama 70B\_q.}
\label{}
\end{table*}

\begin{figure*}[t]
\centering
\begin{tcolorbox}[colback=green!10!white, colframe=green!50!black, title=IHRA Definition with contemporary examples]

{\small ``Antisemitism is a certain perception of Jews, which may be expressed as hatred toward Jews. Rhetorical and physical manifestations of antisemitism are directed toward Jewish or non-Jewish individuals and/or their property, toward Jewish community institutions and religious facilities.'' 
\newline\newline
To guide IHRA in its work, the following examples may serve as illustrations: Manifestations might include the targeting of the state of Israel, conceived as a Jewish collectivity. However, criticism of Israel similar to that leveled against any other country cannot be regarded as antisemitic. Antisemitism frequently charges Jews with conspiring to harm humanity, and it is often used to blame Jews for ``why things go wrong.'' It is expressed in speech, writing, visual forms and action, and employs sinister stereotypes and negative character traits.
\newline\newline
Contemporary examples of antisemitism in public life, the media, schools, the workplace, and in the religious sphere could, taking into account the overall context, include, but are not limited to:
\newline\newline
* Calling for, aiding, or justifying the killing or harming of Jews in the name of a radical ideology or an extremist view of religion.
\newline
* Making mendacious, dehumanizing, demonizing, or stereotypical allegations about Jews as such or the power of Jews as collective — such as, especially but not exclusively, the myth about a world Jewish conspiracy or of Jews controlling the media, economy, government or other societal institutions.
\newline
* Accusing Jews as a people of being responsible for real or imagined wrongdoing committed by a single Jewish person or group, or even for acts committed by non-Jews.
\newline
* Denying the fact, scope, mechanisms (e.g. gas chambers) or intentionality of the genocide of the Jewish people at the hands of National Socialist Germany and its supporters and accomplices during World War II (the Holocaust).
\newline
* Accusing the Jews as a people, or Israel as a state, of inventing or exaggerating the Holocaust.
\newline
* Accusing Jewish citizens of being more loyal to Israel, or to the alleged priorities of Jews worldwide, than to the interests of their own nations.
\newline
* Denying the Jewish people their right to self-determination, e.g., by claiming that the existence of a State of Israel is a racist endeavor.
\newline
* Applying double standards by requiring of it a behavior not expected or demanded of any other democratic nation.
\newline
* Using the symbols and images associated with classic antisemitism (e.g., claims of Jews killing Jesus or blood libel) to characterize Israel or Israelis.
\newline
* Drawing comparisons of contemporary Israeli policy to that of the Nazis.
\newline
* Holding Jews collectively responsible for actions of the state of Israel.
\newline\newline
Antisemitic acts are criminal when they are so defined by law (for example, denial of the Holocaust or distribution of antisemitic materials in some countries). Criminal acts are antisemitic when the targets of attacks, whether they are people or property – such as buildings, schools, places of worship and cemeteries – are selected because they are, or are perceived to be, Jewish or linked to Jews. Antisemitic discrimination is the denial to Jews of opportunities or services available to others and is illegal in many countries.}
\end{tcolorbox}
\caption{IHRA's working definition with contemporary examples~\citet{ihra-def}.}
\label{fig:ihra-definition}
\end{figure*}

\begin{figure*}[t]
\centering
\begin{tcolorbox}[colback=green!10!white, colframe=green!50!black, title=ZS-(AS) Prompt]

\textbf{Role: User}

  \begin{tcolorbox}[colback=white, colframe=black]

  Classify the following Twitter post as antisemitic:
  \newline\newline
  \{\{post\}\}
  \newline\newline
  \#\# At the very end of your response, add following:
  \newline
  - First line must give a summary of your classified decision ``Summary:''
  \newline
  - Second line must give ``Antisemitic:'' (Yes|No)

  \end{tcolorbox}

\end{tcolorbox}
\caption{ZS-(AS) Prompt. The {{post}} is replaced with the actual social media post.}
\label{fig:baseline-prompt}
\end{figure*}






\begin{figure*}[t]
   \centering
   \begin{tcolorbox}[colback=green!10!white, colframe=green!50!black, title=Zero-Shot-{$\alpha$} Prompt]
   
   \textbf{Role: System}
   
      \begin{tcolorbox}[colback=white, colframe=black]
         {\small IHRA Definition: ``Antisemitism is a certain perception of Jews, which may be expressed as hatred toward Jews. Rhetorical and physical manifestations of antisemitism are directed toward Jewish or non-Jewish individuals and/or their property, toward Jewish community institutions and religious facilities.''}
      
      \end{tcolorbox}

   \textbf{Role: User}
   
     \begin{tcolorbox}[colback=white, colframe=black]
      {\small 
     Classify the following Twitter post as antisemitic based on IHRA's definitions:
     \newline\newline
     \{\{post\}\}
     \newline\newline
     \#\# At the very end of your response, add following:
     \newline
     - First line must give a summary of your classified decision ``Summary:''
     \newline
     - Second line must give ``Antisemitic:'' (Yes|No)
      }
     \end{tcolorbox}
   
   \end{tcolorbox}
   \caption{Zero-Shot-{$\alpha$} Prompt. In this prompting, we only provide IHRA definition and do not provide contemporary examples.}
   \label{fig:zeroshot-alpha-prompt}
\end{figure*}


\begin{figure*}[t]
   \centering
   \begin{tcolorbox}[colback=green!10!white, colframe=green!50!black, title=Zero-Shot-{$\beta$} Prompt]
   
   \textbf{Role: System}
   
      \begin{tcolorbox}[colback=white, colframe=black]
         {\small \{\{IHRA Definition with contemporary examples\}\}}
      \end{tcolorbox}

   \textbf{Role: User}
   
     \begin{tcolorbox}[colback=white, colframe=black]
      {\small 
     Classify the following Twitter post as antisemitic based on IHRA's definitions:
     \newline\newline
     \{\{post\}\}
     \newline\newline
     \#\# At the very end of your response, add following:
     \newline
     - First line must give a summary of your classified decision ``Summary:''
     \newline
     - Second line must give ``Antisemitic:'' (Yes|No)
      }
     \end{tcolorbox}
   
   \end{tcolorbox}
   \caption{Zero-Shot-{$\beta$} Prompt. In this prompting, we provide a complete IHRA definition with contemporary examples and surrounding text on the IHRA's website (see Fig.~\ref{fig:ihra-definition}, ~\citet{ihra-def}).}
   \label{fig:zeroshot-beta-prompt}
   
\end{figure*}

\begin{figure*}[t]
   \centering
   \begin{tcolorbox}[colback=green!10!white, colframe=green!50!black, title=Zero-Shot-CoT Prompt]
   
   \textbf{Role: System}
   
      \begin{tcolorbox}[colback=white, colframe=black]
         {\small \{\{IHRA Definition with contemporary examples\}\}}
      \end{tcolorbox}

   \textbf{Role: User}
   
     \begin{tcolorbox}[colback=white, colframe=black]
      {\small 
      Classify the following Twitter post as antisemitic based on IHRA's definitions:
     \newline\newline
     \{\{post\}\}
     \newline\newline
     \#\# Let's think step by step. And at the very end of your response, add following:
     \newline
     - First line must give a summary of your classified decision ``Summary:''
     \newline
     - Second line must give ``Antisemitic:'' (Yes|No)
      }
     \end{tcolorbox}
   
   \end{tcolorbox}
   \caption{Zero-Shot-{$\beta$} Prompt. In this prompting, we provide a complete IHRA definition with contemporary examples and surrounding text on the IHRA's website}
   \label{fig:zeroshot-cot-prompt}

\end{figure*}

\begin{figure*}[t]
   \centering
   \begin{tcolorbox}[colback=blue!10!white, colframe=blue!50!black, title=Guided-CoT Prompt]
   
   \textbf{Role: System}
   
      \begin{tcolorbox}[colback=white, colframe=black]
         {\small \{\{IHRA Definition with contemporary examples\}\}}
      \end{tcolorbox}

   \textbf{Role: User}
   
     \begin{tcolorbox}[colback=white, colframe=black]
      {\small 
      Classify the following Twitter post as antisemitic based on IHRA's definitions:
     \newline\newline
     \{\{post\}\}
     \newline\newline
     \#\# Let's analyze and think step by step:
     \newline
     1. Break the post into chunks and explain what is being conveyed
     \newline
     2. Identify whether OP's post is sarcasm, criticism, quoted statements, news, coded language, ambiguous phrasing, or reporting etc.?
     \newline
     3. Check if OP is calling out to promote antisemitism or using statements/narrative to spread awareness about antisemitism
     \newline
     4. Assess whether the post potentially aligns with the IHRA definition of antisemitism and its contemporary examples
     \newline
     5. Based on your analysis, decide whether OP's stance is favorable to the Jews community or not?
       \newline\newline
       \#\# And at the very end of your response, add following:
       \newline
       - First line must give a summary of your classified decision ``Summary:''
       \newline
       - Second line must give ``Antisemitic:'' (Yes|No)
      }
     \end{tcolorbox}
   \end{tcolorbox}
   \caption{Guided-CoT prompt (ours).}
   \label{fig:guided-cot-prompt}
\end{figure*}

\end{document}